\documentclass[runningheads]{llncs}

\usepackage{eccv}

\usepackage{eccvabbrv}

\usepackage{graphicx}
\usepackage{booktabs}
\usepackage{multirow}
\newcommand{\nzk}[1]{{\color{black}#1}}

\usepackage[accsupp]{axessibility}  

\usepackage{hyperref}

\usepackage{orcidlink}

\begin{document}

\title{SV2V-RSim: A Comprehensive Benchmark for Self-Selective V2V Cooperative
Perception with Near-Realistic Data} 

\titlerunning{Abbreviated paper title}

\author{
Yulu Wu\inst{1} \and
Chao Wei\inst{1} \and
Jujun Cheng\inst{1} \and
Zhangkai Ni\inst{1} \and
Haowen Wang\inst{2} \and
Dengyang Suo\inst{1} \and
Cong Chen\inst{1} \and
Xinyi Liu\inst{1} \and
Shangce Gao\inst{3}
}

\authorrunning{Y.~Wu et al.}

\institute{
Tongji University, Shanghai, China
\and
Anhui University, Hefei, China
\and
University of Toyama, Toyama, Japan
}

\maketitle

\begin{abstract}
  Vehicle-to-Vehicle (V2V) cooperative perception enhances
autonomous driving by enabling vehicles to share information beyond their direct line of sight. However, existing V2V
datasets are limited by a small number of participating agents,
static collaborator selection strategies, and a significant domain gap between simulated and real-world environments.
To overcome these challenges, we introduce SV2V-RSim—a
large-scale, multi-modal, near-realistic simulation dataset engineered to elevate agent diversity and realism. Additionally,
we present the Select Vehicles Adaptively (SVA) module,
which optimizes collaborator selection to balance perception
performance against communication bandwidth constraints.
Our dataset is generated using the Unreal Engine 5-based
simulator that integrates high-fidelity 3D assets, diverse environments, and intricate traffic scenarios. All vehicles within
a specified range of the ego vehicle are equipped with sensor suites, enabling dynamic and adaptive collaborator selection. SV2V-RSim encompasses four maps, four weather conditions, six time periods from sunrise to night, 203K LiDAR
frames, 402K RGB frames, and 788K annotated 3D bounding
boxes across 17 object classes, supporting a range of cooperative perception tasks such as 3D object detection, segmentation, and depth estimation. Benchmarking on recent cooperative perception algorithms demonstrates that SVA achieves a
superior performance-bandwidth trade-off, while sim-to-real
experiments and No-Reference Image Quality Assessment
validate the dataset’s high realism and practical effectiveness.
Our dataset and code will be publicly available.
  \keywords{Collaborative Perception\and Autonomous Driving\and Synthetic Data}
\end{abstract}

\section{Introduction}
\label{sec:intro}

Perception is the vital sentinel that underpins both safety and intelligent decision-making in autonomous vehicles. 
\nzk{Recent advances in deep learning and the development of high-quality autonomous driving datasets have significantly advanced single-vehicle perception tasks, such as 3D object detection~\cite{mao20233d} and semantic segmentation~\cite{di2021seg}, achieving remarkable improvements in accuracy.}
However, within the complexity of open-world environments, elusive corner cases and occlusion-related challenges continue to reveal the intrinsic limitations of single-vehicle perception~\cite{liu2023v2xsurvey}.
\nzk{To overcome these limitations arising from constrained viewpoints, Vehicle-to-Vehicle (V2V) cooperative perception has emerged as a promising paradigm~\cite{wang2020v2vnet}. }
By facilitating the exchange of sensory information between autonomous vehicles, V2V cooperative perception enhances the ego vehicle’s situational awareness, thereby effectively mitigating safety issues associated with occlusions and limited sensing ranges.

\begin{figure}[t]
\centering
    \centering{\includegraphics[width=1.0\linewidth]{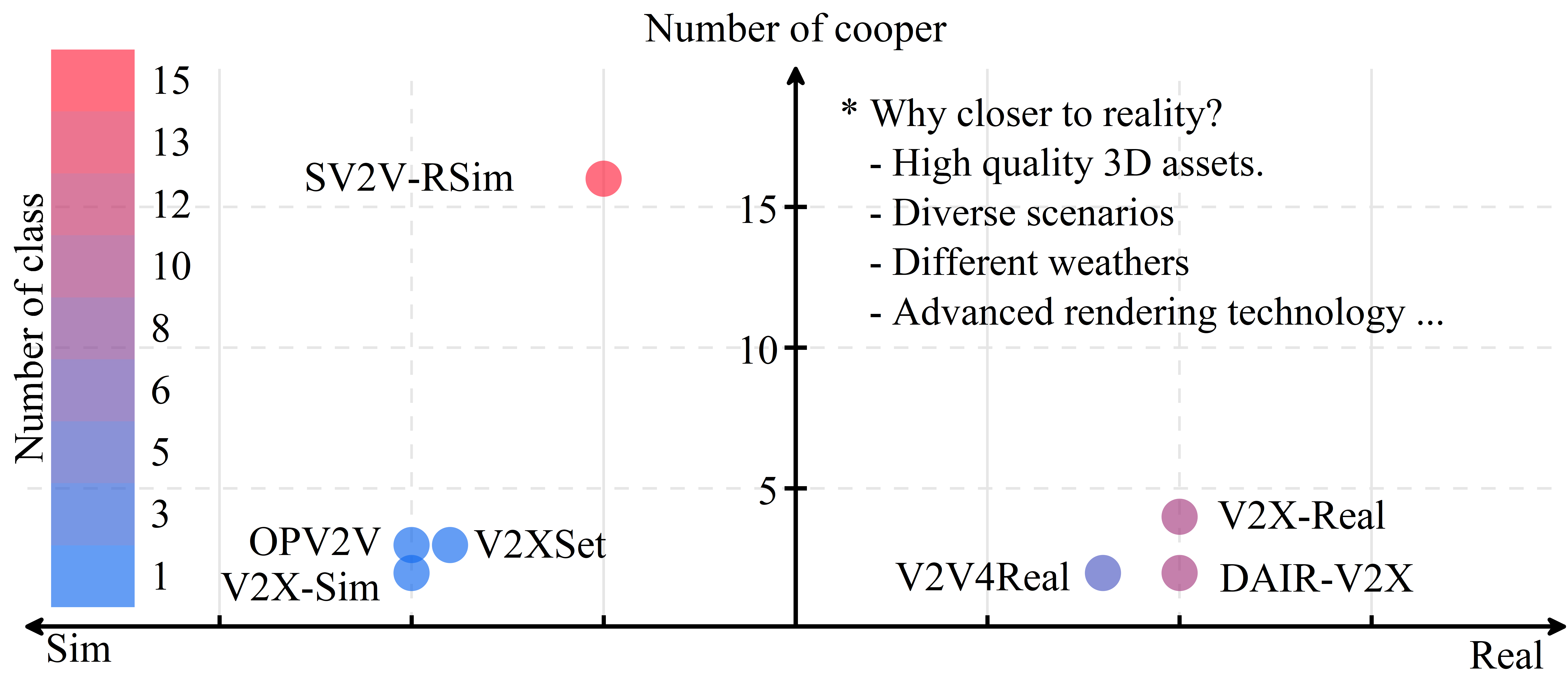}}
\caption{Datasets for cooperative perception. The color represents the number of annotation categories. SV2V-RSim has a wide range of collaborators to choose from. The closer the horizontal axis of the simulation data is to 0, the smaller the gap between it and the real data. We use a series of methods to enhance the realism and generalization of the dataset.}
\label{fig:datasets}
\end{figure}

Driven by the advantages of cooperative perception, many researchers have focused on enhancing its accuracy~\cite{wang2020v2vnet,xu2022v2x,xu2022cobevt,wang2024iftr} and communication efficiency~\cite{liu2020who2comm,hu2022where2comm}. In addition, recent studies address real-world challenges such as sensor model discrepancies~\cite{li2024s2r} and pose errors~\cite{vadivelu2021learning}. However, existing cooperative perception datasets still exhibit notable limitations.
Real-world datasets, including DAIR-V2X~\cite{yu2022dair-v2x}, V2V4Real~\cite{xu2023v2v4real}, V2XReal~\cite{xiang2024v2xreal}, RCooper~\cite{hao2024rcooper}, and UrbanIng-V2X~\cite{sekaran2026urbaningv2xlargescalemultivehiclemultiinfrastructure}, are constrained by the prohibitive costs associated with deploying a sufficient number of cooperative agents, thereby hindering a comprehensive understanding of the surrounding environment. 
\nzk{In contrast, simulation datasets such as OPV2V~\cite{xu2022opv2v}, V2XSet~\cite{xu2022v2x}, and V2X-Sim~\cite{li2022v2xsim} leverage CARLA-SUMO co-simulation to generate customizable environments, yet suffer from fixed collaborator selection schemes that hinder adaptation to dynamic scenarios.
Moreover, the acquisition of high-fidelity 3D assets and inconsistencies in near-realistic rendering further widen the sim-to-real gap.}

\begin{table}[!t]
\centering
\scriptsize
\setlength{\tabcolsep}{1.4pt}
\begin{tabular}{l|l|c|c|c|c|c|c|c|c}
\toprule
Dataset &
  Year &
  \multicolumn{1}{c|}{\begin{tabular}[c]{@{}c@{}}Real/\\ Sim\end{tabular}} &
  V2X &
  \multicolumn{1}{c|}{\begin{tabular}[c]{@{}c@{}} Avg Number \\
  of Agents\end{tabular}} &
  \begin{tabular}[c]{@{}l@{}}RGB\\ images\end{tabular} &
  \multicolumn{1}{c|}{LiDAR} &
  \multicolumn{1}{c|}{\begin{tabular}[c]{@{}c@{}}3D\\ boxes\end{tabular}} &
  \multicolumn{1}{c|}{Classes} &
  Locations \\ \midrule
Kitti &
  2012 &
  Real &
  No &
  1 &
  15k &
  15k &
  200k &
  8 &
  Karlsruhe \\
nuScenes &
  2019 &
  Real &
  No &
  1 &
  1.4M &
  400k &
  1.4M &
  23 &
  Boston, SG \\
Waymo Open &
  2019 &
  Real &
  No &
  1 &
  1M &
  200k &
  12M &
  4 &
  3x USA \\
OPV2V &
  2022 &
  Sim &
  V2V &
  3 &
  44k &
  11k &
  230k &
  1 &
  \begin{tabular}[c]{@{}c@{}}CARLA \end{tabular} \\
V2X-Sim &
  2022 &
  Sim &
  \begin{tabular}[c]{@{}l@{}}V2V\&I\ \end{tabular} &
  2 &
  60K &
  10k &
  26.6k &
  1 &
  \begin{tabular}[c]{@{}c@{}}CARLA\end{tabular} \\
V2XSet &
  2022 &
  Sim &
  \begin{tabular}[c]{@{}l@{}} V2V\&I\end{tabular} &
  3 &
  44K &
  11k &
  230k &
  1 &
  \begin{tabular}[c]{@{}c@{}}CARLA\end{tabular} \\
DAIR-V2X &
  2022 &
  Real &
  V2I &
  2 &
  39K &
  39K &
  464K &
  10 &
  Beijing, CN \\ 
V2V4Real &
  2022 &
  Real &
  V2V &
  2 &
  40K &
  20K &
  240K &
  5 &
  \begin{tabular}[c]{@{}c@{}} Ohio, USA\end{tabular} \\ 
  
V2XReal &
  2024 &
  Real &
  \begin{tabular}[c]{@{}l@{}} V2V\&I\end{tabular} &
  4 &
  171K &
  33K &
  1.2M &
  10 &
  \begin{tabular}[c]{@{}c@{}} Ohio, USA\end{tabular} \\

  UrbanIng-V2X &
  2025 &
  Real &
  \begin{tabular}[c]{@{}l@{}} V2V\&I\end{tabular} &
  3 &
  81.6k &
  27.2k &
  712k &
  13 &
  \begin{tabular}[c]{@{}c@{}} Ingolstadt, Germany\end{tabular} \\\midrule
SV2V-RSim (ours) &
  2026 &
  Sim &
  V2V &
  16.2 &
  402K &
  203K &
  788K &
  17 &
  \begin{tabular}[c]{@{}c@{}} Unreal Engine 5\\-based simulator
  \end{tabular}
  \\\bottomrule
\end{tabular}
\caption{Comparison between the proposed dataset and other notable autonomous driving datasets.}
\label{tab:datasets}
\end{table}

Data is collected using the Unreal Engine 5-based simulator equipped with highly generalized 3D assets, complex traffic scenarios, and diverse environmental systems. 
\nzk{To overcome these limitations, particularly the lack of dynamic collaborator selection and restricted realism, we propose SV2V-RSim, a large-scale, multi-modal, multi-task simulation dataset for V2V cooperative perception. 
SV2V-RSim is the first dataset to enable dynamic collaborator selection over time, capturing the fluid and context-dependent nature of real-world driving scenarios.
Built on an Unreal Engine 5-based simulator, SV2V-RSim integrates high-quality 3D assets, procedurally generated environments, diverse traffic patterns, and varied rendering conditions to improve both realism and generalization. }
At each timestamp, every autonomous vehicle within a 70-meter radius of the ego vehicle is outfitted with a comprehensive sensor suite, ensuring the collection of rich and varied perception data. The SV2V-RSim dataset spans four maps, incorporates four distinct weather conditions and six time periods from sunrise to night, and comprises 203K LiDAR frames, 402K RGB frames, and 788K annotated 3D bounding boxes across 17 object classes.
These intricate environments, marked by a high density of road users and constantly evolving traffic patterns, offer a new benchmark dataset for advancing V2V cooperative perception research.

Balancing performance and bandwidth becomes increasingly challenging as more agents fall within the communication range. 
Moreover, the fluid nature of collaborative environments compels autonomous vehicles to continuously reselect their interaction targets in real time. To address these issues, we introduce the Select Vehicles Adaptively (SVA) module—a dynamic framework that leverages real-time sensor data and contextual awareness to intelligently identify the most informative collaborators. 
Our contributions can be summarized as follows:
\begin{itemize}
    \item 
    \nzk{We build SV2V-RSim, the first large-scale simulation dataset for V2V cooperative perception that incorporates dynamic collaborator selection. 
    Built using Unreal Engine 5, it uses high-fidelity assets, realistic scenes, procedural content generation, and diverse rendering conditions to improve both realism and scalability.}
    
    \item 
    To handle collaborative scenarios with dynamic, real-time changes, we propose the Select Vehicles Adaptively (SVA) module, which balances detection performance and bandwidth efficiency by leveraging low-level cooperative vehicle data to prioritize task-critical features for ego perception. This resource-aware and contextually adaptive collaborative sensing strategy optimizes communication and perceptual outcomes.
    
    \item 
    \nzk{SV2V-RSim supports a variety of cooperative tasks. We benchmark several state-of-the-art cooperative perception methods using SV2V-RSim and SVA, and further validate the dataset’s realism and generalization through sim-to-real experiments.}
 \end{itemize}

\section{Related work}
\label{sec:related_work}
\textbf{Simulator for autonomous driving.} Simulators play a crucial role in generating autonomous driving simulation data~\cite{li2024choose}. Early public simulators, such as CARLA~\cite{dosovitskiy2017carla} and AirSim~\cite{airsim2017fsr}, use their built-in maps and vehicle models to generate customized driving scenarios for data collecting. However, early simulators lacked sufficiently high‑fidelity asset models and rendering capabilities, resulting in a discrepancy between simulated and real‑world data. To better simulate complex traffic flows, SUMO~\cite{krajzewicz2012sumo} is often used for joint simulation. OpenCDA~\cite{xu2021opencda} is an open CARLA-SUMO co-simulation for cooperative driving. 
\nzk{However, OpenCDA inherits key limitations from CARLA, including the challenge of balancing rendering quality, agent count, and hardware constraints.}

\textbf{Autonomous Driving Datasets.} High-quality public datasets have provided tremendous support for the development of autonomous driving. Table \ref{tab:datasets} shows some recent autonomous driving datasets and their differences. Kitti~\cite{Geiger2012kitti} is the first autonomous driving dataset with 3D labels, providing images, LiDAR points, and GPS data. The nuScenes~\cite{caesar2020nuscenes} and Waymo Open dataset~\cite{Sun2020waymo} are two of the most widely used large-scale datasets in autonomous driving recently. However, in complex and urgent scenarios, single-vehicle perception struggles to cope.

V2V cooperative perception technology expands the perception range and provides more diverse perspectives by sharing and fusing multi-modal sensory information with surrounding vehicles. OPV2V~\cite{xu2022opv2v} is the first V2V cooperative perception dataset. Compared to OPV2V, V2X-Sim~\cite{li2022v2xsim} and V2XSet~\cite{xu2022v2x} incorporate information sharing between vehicles and roadside infrastructure (V2I), enriching cooperative perception with a broader range of collaborative sources. Compared to simulation, real-world data more accurately reflects actual driving scenarios. DAIR-V2X~\cite{yu2022dair-v2x}, V2V4Real~\cite{xu2023v2v4real} and V2XReal~\cite{xiang2024v2xreal} provide high-quality collaborative data. 
However, due to the high cost of data collection, these real-world datasets involve only a limited number of collaborators, limiting their ability to fully showcase the benefits of cooperative perception.

\textbf{Cooperative Perception.} Cooperative perception technology enhances the perception capability and range of autonomous vehicles by sharing and fusing information among different collaborators. V2V cooperative perception focuses on information exchange between vehicles~\cite{liu2023v2xsurvey}. Existing methods typically adopt early, intermediate, or late fusion strategies~\cite{liu2023v2xsurvey}.  V2VNet~\cite{wang2020v2vnet} uses graph neural networks to fuse features. V2X-ViT~\cite{xu2022v2x} uses a unified vision transformer to achieve robust collaborative perception, while CoBEVT~\cite{xu2022cobevt} introduces a flexible transformer framework designed for similar applications. Furthermore, to better balance communication bandwidth and perception performance, Where2comm~\cite{hu2022where2comm} and How2comm~\cite{yang2023how2comm} incorporate a spatial confidence map to represent spatial diversity within perceptual data. 
This approach partially filters collaborators and collaborative information, but its effectiveness is difficult to prove when collaborators are changing.
\nzk{In this paper, we validate the proposed benchmark method on SV2V-RSim through extensive experiments, demonstrating its effectiveness across multiple tasks.}

\begin{figure*}[!t]
\centering
    \centering{\includegraphics[width=1\linewidth]{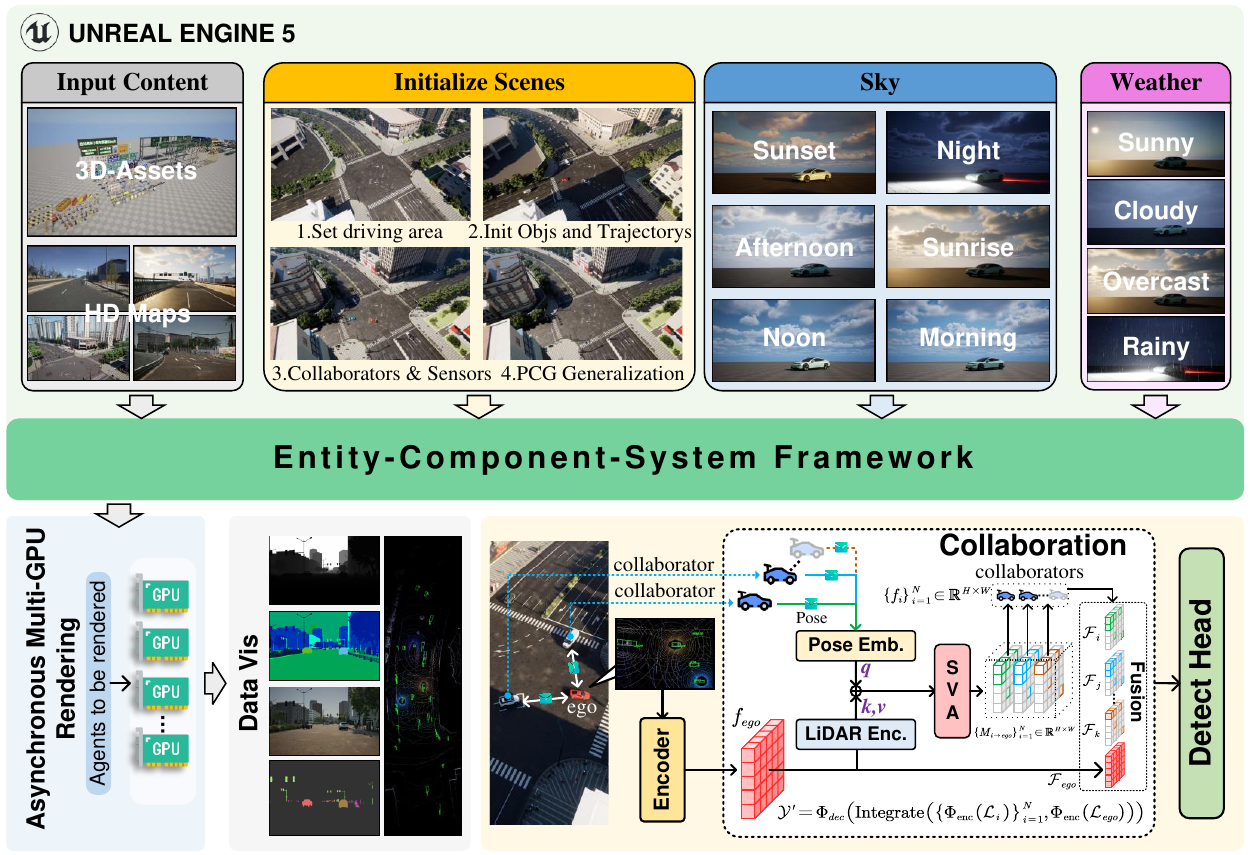}}
\caption{Pipeline of data production and self-selective cooperative perception network.}
\label{fig:pipeline}
\end{figure*}

\section{SV2V-RSim Dataset}

To address real-world challenges in collaborative perception, we propose SV2V-RSim, a large-scale, multi-modal, near-realistic dynamic simulation dataset for V2V cooperative perception. The dataset is annotated with 3D bounding boxes, depth labels, and both semantic and instance segmentation labels across 17 categories.
Figure~\ref{fig:pipeline} illustrates the data collection pipeline and perception methodology. 
\nzk{We begin by detailing the simulation design, followed by the setup and collection of the dataset. Lastly, we conduct a comprehensive data analysis. A demonstration video is provided in the supplementary materials.}

\subsection{Unreal Engine 5-based Simulation}
\label{sec:ue_sim}
Unreal Engine (UE) has emerged as a cornerstone in autonomous driving simulation, celebrated for its high-quality graphics rendering and realistic physics simulation~\cite{ue4}. Compared to UE4, UE5~\cite{ue5} brings significant advancements with Nanite and Lumen, which enable real-time global illumination and excellent geometric detail. To further enhance realism, it is essential to integrate high-quality 3D assets, diverse HD maps, and dynamic weather and sky systems. Moreover, improving the rendering framework itself is equally pivotal in elevating V2V simulation data quality. 

\nzk{Creating a near-realistic simulation dataset requires improvements across all simulation components.}
Our goal is to improve the performance of each module individually, while also ensuring their coupling to generate high-quality data. Additionally, to better utilize the existing assets, we have performed local development on UE5 based on annotations and operations. Excluding differences in assets and maps, our method provides a more efficient and generalizable solution compared to existing simulation approaches. \nzk{In the following sections, we provide a detailed introduction to these modules.}

\begin{figure}[t]
    \centering
    \begin{subfigure}{0.21\textwidth}
        \centering
        \includegraphics[width=\textwidth]{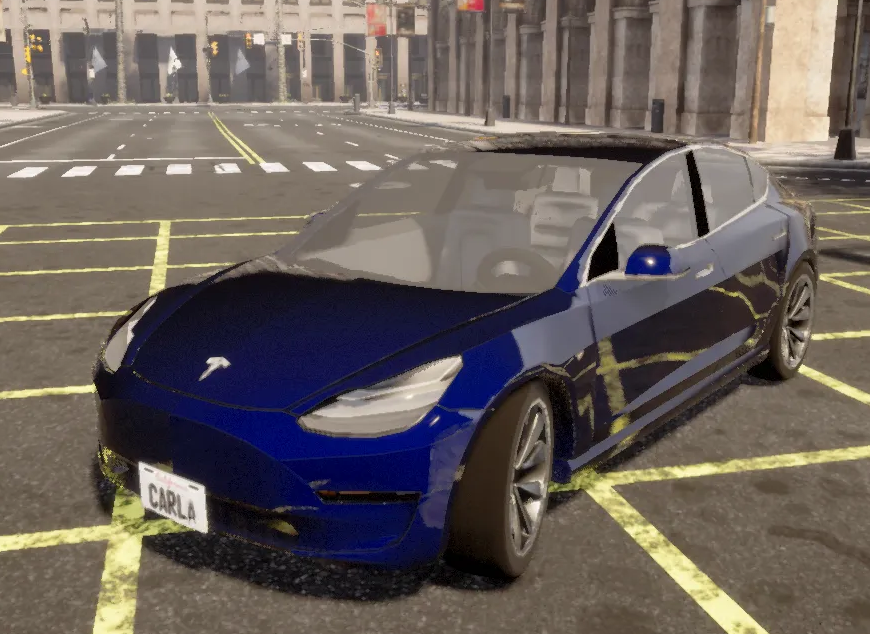} 
        \caption{Carla's Asset}
    \end{subfigure}
    \hspace{0.5cm} 
    \begin{subfigure}{0.21\textwidth}
        \centering
        \includegraphics[width=\textwidth]{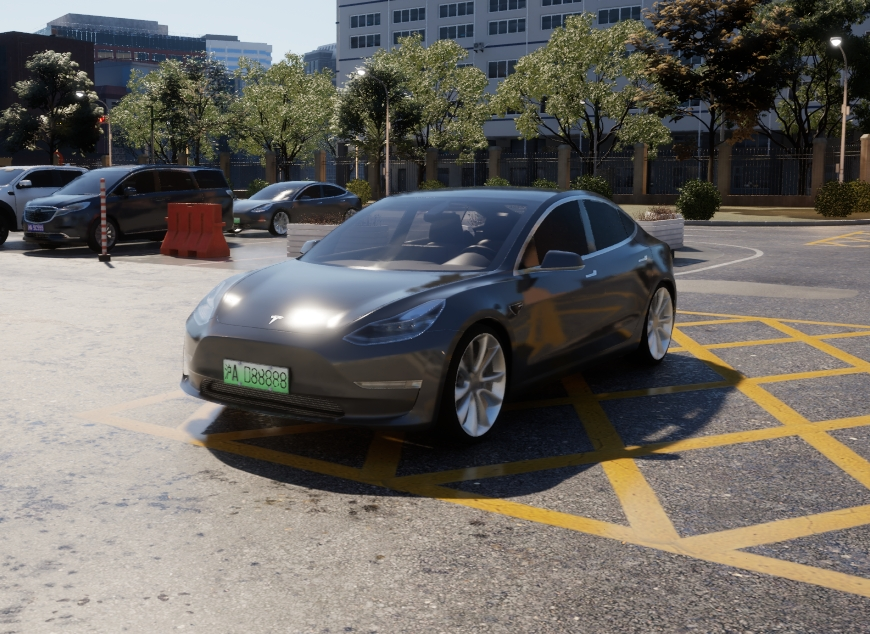} 
        \caption{Our Asset}
    \end{subfigure}
    \caption{Comparison between Tesla Model 3 assets in CARLA and SV2V-RSim.}
    \label{fig:comparison_asset}
\end{figure}

\subsubsection{3D assets} 
High-quality 3D assets serve as a vital bridge in narrowing the gap between simulation and reality~\cite{liu2024vqa}. \nzk{We construct an asset library by aggregating open-source assets and generating a set of high-quality assets de novo.} 
\nzk{Compared to those used in existing simulation datasets, our assets offer finer geometric detail and more realistic material properties. 
As shown in Figure~\ref{fig:comparison_asset}, our Tesla Model 3 asset provides a substantial improvement over the one in CARLA. More examples are provided in the supplementary materials.
We categorize our assets into four groups: 91 motor vehicles, 9 non-motor vehicles, 10 static object, and 17 human types. 
To further improve generalizability, each asset's material is randomized via Procedural Content Generation (PCG)~\cite{liu2021pcg}.}

\subsubsection{HD Maps} 
\nzk{To reflect the diversity of real-world driving, our simulation includes HD maps representing urban centers, suburban roads, and highways.}
These maps feature routes that vary from simple straight paths to complex networks with intersections, highway forks, roundabouts, and dedicated entry and exit sections. 
To further enrich environmental diversity and improve domain generalization, we employ PCG techniques. Rendered images of these HD maps are provided in the supplementary material.

\subsubsection{Weather and Sky Systems} 
\nzk{Weather conditions and time of day significantly influence perception~\cite{mușat2021multi}. 
Therefore, we incorporate an advanced weather and sky system that supports four weather types—sunny, cloudy, overcast, and rainy—and spans six daily time periods from sunrise to night.}
Moreover, these systems are tightly coupled with the vehicle system; for example, headlights will be activated automatically under low ambient light conditions. Rendered images illustrating these diverse conditions are provided in the supplementary material.

\subsubsection{Rendering framework} 
To achieve high-quality rendering at scale, we design an asynchronous multi-GPU rendering framework. Each GPU is assigned a dedicated renderer that communicates with the simulator via a unique port, ensuring an even distribution of sensor rendering tasks. This parallelization not only accelerates the rendering process but also maintains the temporal coherence of diverse outputs—images, point clouds, annotations, and more—through inter-renderer communication.

\subsection{Data Setup and Collection}
\label{sec:data_acquisition}

\subsubsection{Sensor Setup}
All vehicles are equipped with a comprehensive sensor suite, including cameras, LiDAR, GPS, and IMU. To mitigate the high computational costs and potential degradation in rendering quality associated with simultaneous sensor activation, we implement a conditional sensor activation strategy. 
Specifically, sensors are activated only when a vehicle is within a 70-meter radius of the ego vehicle, an effective communication range as established in prior work~\cite{xu2022v2x}.
The sensor configuration is based on the nuScenes setup~\cite{caesar2020nuscenes}. 
Each vehicle is equipped with both front and rear cameras, which capture RGB, depth, and semantic segmentation data.  
Additionally, a LiDAR sensor is mounted on the top of the vehicle. Detailed sensor specifications are provided in \nzk{Table} \ref{tab:sensor_config}.

\begin{table}[t]
\small
    \centering
    \begin{tabular}{ll}
        \toprule
        \textbf{Sensors} & \textbf{Details} \\ \midrule
        2x Cameras & RGB, Depth, Semantic \& Instance Seg, \\
                   & 1600 × 900 resolution, 5Hz frequency. \\ 
        1x LiDAR   & Semantic \& Instance Seg, \\ 
                   & 5Hz frequency, $\leq$ 200m range,\\
                   & 32 beams, 1080 points per ring, \\
                   & 32 channels, 360° horizontal FOV, \\ \bottomrule
    \end{tabular}
    \caption{Sensor Configuration}
    \label{tab:sensor_config}
\end{table}

\subsubsection{Scenario Setup} 
\nzk{We design and simulate 210 customized driving scenarios across a variety of road types, including urban streets, suburban roads, highways, and roundabouts. 
Agents in these scenarios exhibit diverse behaviors, such as acceleration, deceleration, lane changes, overtaking, and turning maneuvers. }
Each scenario is simulated for 30 seconds, with sky and weather conditions determined based on predefined probability distributions.

\subsubsection{Collection Workflow} 
After preparing the required 3D assets and scene elements, such as maps, traffic flows, environmental parameters, and procedural generation settings, we equip each vehicle with the full sensor suite and configure conditional activation switches. 
\nzk{Rendering tasks are then distributed across multiple GPUs, each running an independent rendering program.}
When the simulation program is running, the ego vehicle begins its trajectory, triggering the activation of sensors on nearby vehicles within a defined range. 
These vehicles are then randomly distributed among the available renderers for simultaneous processing. 
Finally, by synchronizing the timestamps of all collaborating annotators, the complete dataset is obtained.

\subsection{Data Analysis}
\label{sec:data_statistics}

\subsubsection{Data Statistics}
\label{sec:data_statistics_analysis}

\begin{figure}[t]
  \centering
  \includegraphics[width=3.2in]{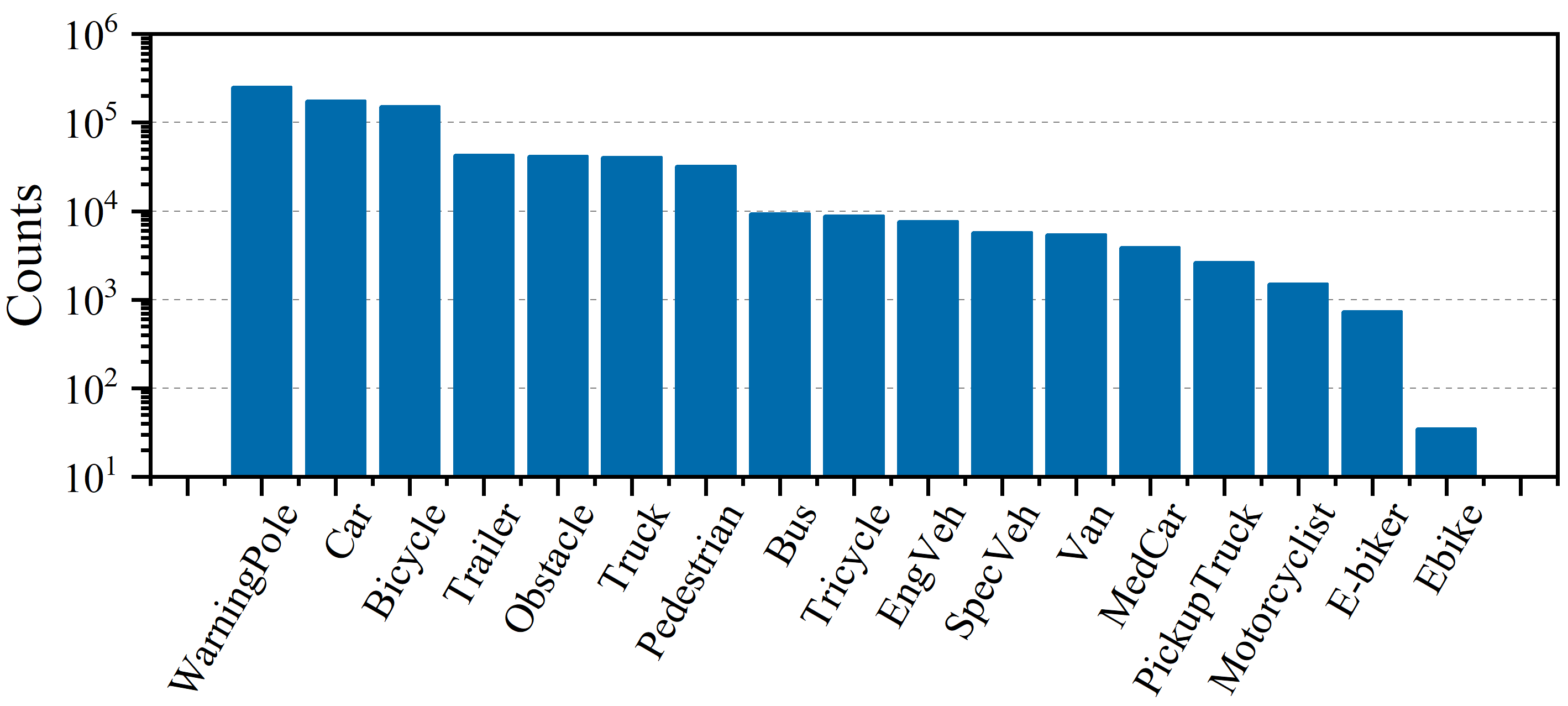}
   \caption{Number of annotations per category.}
   \label{fig:annotation_number}
\end{figure}

\nzk{SV2V-RSim comprises 203K LiDAR frames, 402K RGB images, and 788K annotated 3D bounding boxes across 17 object classes. 
The annotation count per category is illustrated in Figure~\ref{fig:annotation_number}. 
Compared to existing datasets, SV2V-RSim features a broader range of vehicle types and provides denser annotations for small objects such as static obstacles and bicycles.
To enhance realism and diversity, scenarios vary in road types, weather conditions, asset models, and vehicle motion patterns, addressing the complexities of real-world driving environments. 
On average, each scenario involves 16.2 collaborative agents, offering significantly more collaborative opportunities than existing datasets.}

\begin{figure}[htbp]
    \centering
    \begin{minipage}{0.34\textwidth}
        \centering
        \includegraphics[width=\linewidth]{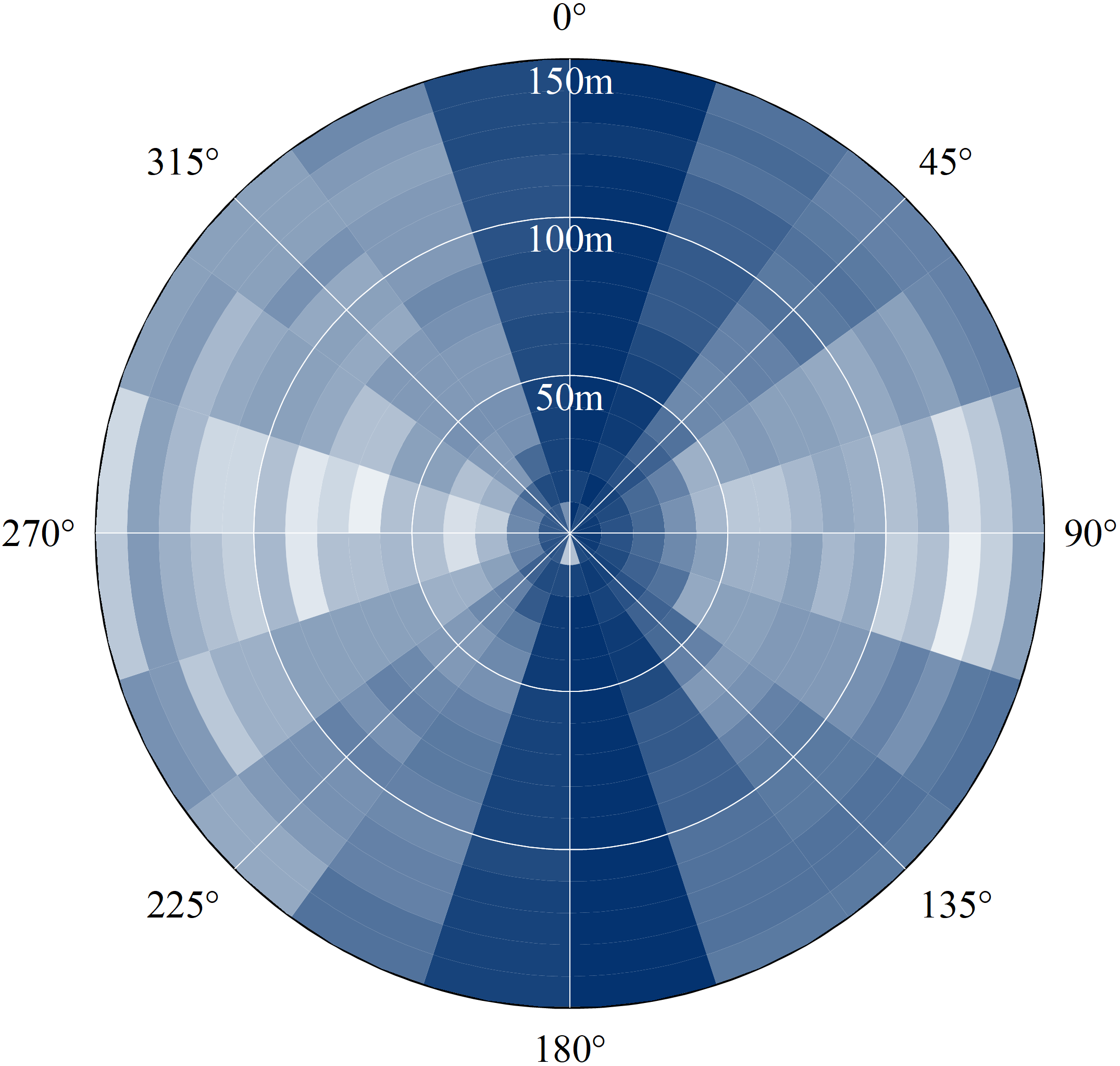}
        \subcaption{Motor}
    \end{minipage}\hspace{0.01cm} 
    \begin{minipage}{0.34\textwidth}
        \centering
        \includegraphics[width=\linewidth]{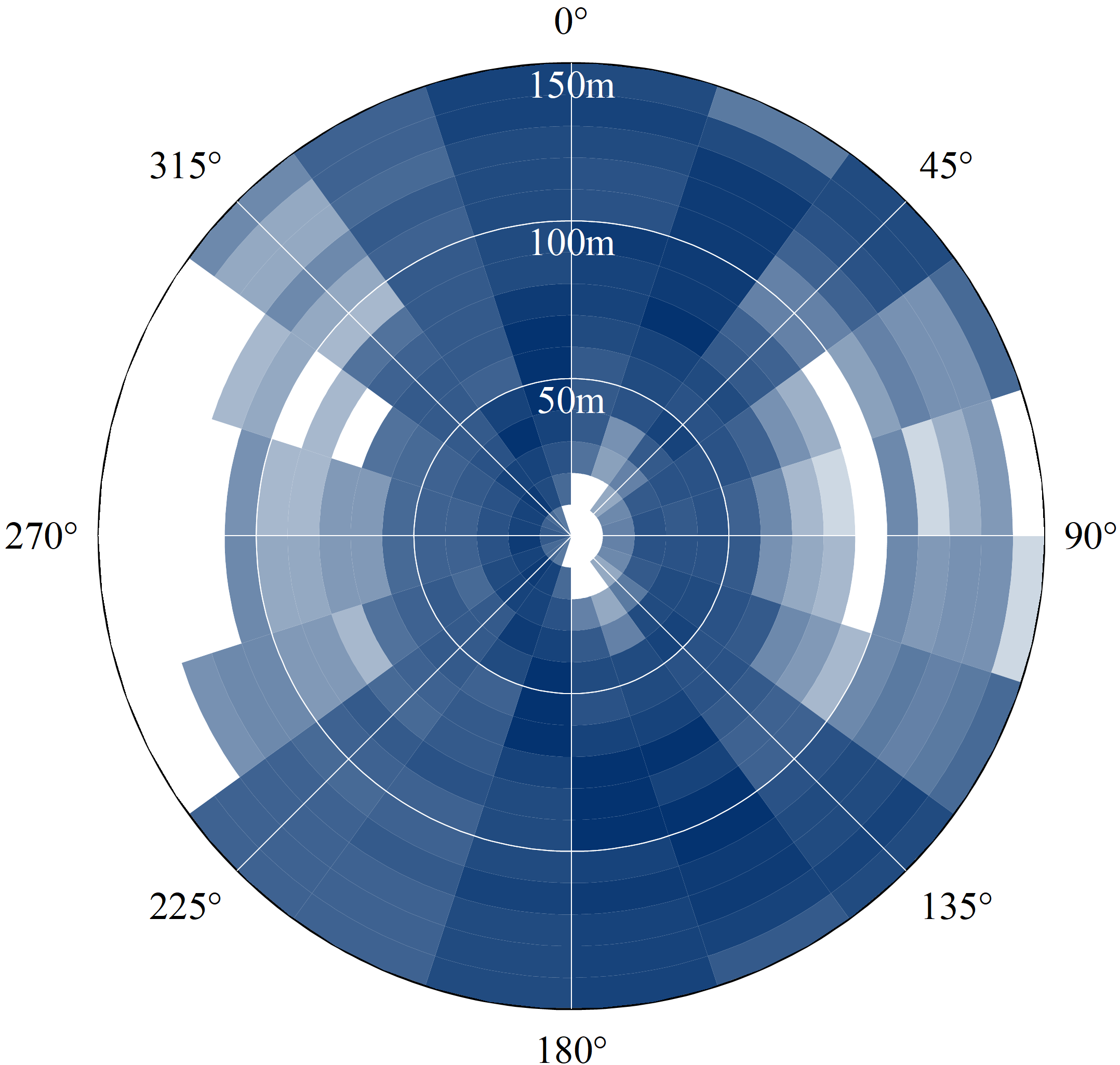}
        \subcaption{Non-Motor}
    \end{minipage}\hspace{0.01cm} 
    \caption{Polar log-scaled density map of box annotations for Motor and Non-Motor, where the radial axis represents the distance, and the angular axis corresponds to the yaw angle. The darker the color, the higher the annotation count. Results for the other categories are available in the supplementary materials.}
    \label{fig:pos_dis}
\end{figure}

\nzk{We group all object classes into four categories based on their semantic attributes: motor vehicles, non-motor vehicles, pedestrians, and static objects. }
Figure \ref{fig:pos_dis} shows the positional distribution of the motors and non-motors within a 150m radial distance. 
\nzk{Each bin on the polar plot corresponds to a spatial region defined by distance and yaw angle relative to the ego vehicle.}
We can observe that objects are mainly distributed in the front and rear, which is due to the road structure consisting of one-way and two-way lanes. 
At the same time, the distribution angle of people and static objects is larger because they are mostly located near sidewalks along the road. 
Moreover, the inclusion of intersection scenarios allows various agents to cross the ego vehicle’s path, leading to additional annotations appearing on the lateral sides.

\begin{figure}[t]
  \centering
  \includegraphics[width=1\linewidth]{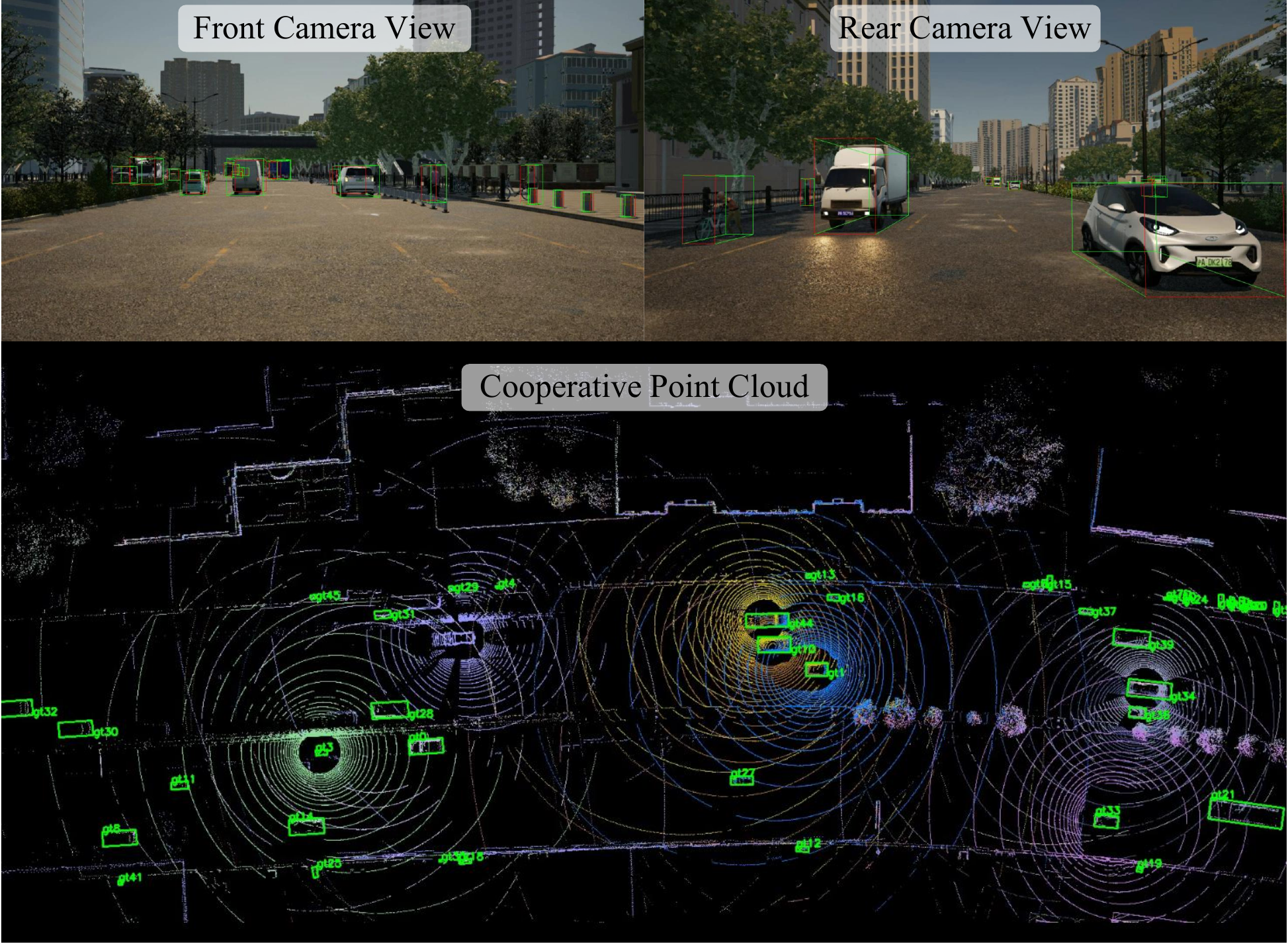}
   \caption{Visualization of a typical multi-agent cooperative scenario in SV2V-RSim. The top row displays the high-fidelity front and rear RGB camera views from the ego vehicle, overlaid with precise 3D bounding box annotations. The bottom row illustrates the cooperative bird's-eye view (BEV) point cloud, which aggregates LiDAR data from the ego vehicle and other collaborative agents within the communication range. The distinct overlapping scanning rings demonstrate how V2V multi-sensor fusion significantly expands the perception field and mitigates severe occlusions.}
   \label{fig:cooperative scenario}
\end{figure}

To intuitively demonstrate the multi-agent nature and sensory richness of SV2V-RSim, Figure \ref{fig:cooperative scenario} visualizes a representative cooperative perception scenario from the perspective of an ego vehicle. The high-fidelity RGB images (Figure \ref{fig:cooperative scenario}, top) showcase the ego vehicle's front and rear camera views, capturing intricate environmental details, realistic lighting, and precise 3D annotations. However, single-vehicle perception is inherently limited by occlusions and a restricted field of view. To overcome this, SV2V-RSim enables dynamic information sharing. As shown in the aggregated cooperative point cloud (Figure \ref{fig:cooperative scenario}, bottom), the ego vehicle seamlessly integrates LiDAR data from surrounding collaborators within its communication range. The overlapping concentric scanning rings provide dense, continuous coverage of the complex traffic environment, establishing a robust foundation for evaluating cooperative 3D object detection algorithms.


\subsubsection{Sim2Real Domain Gap}
\label{sec:data_sim2real}
\nzk{In this section, we evaluate the realism and generalization capability of SV2V-RSim by transferring models trained on synthetic data to real-world scenarios. 
This sim-to-real assessment aims to examine how well our benchmark adapts to the inherent complexities and variations of real-world environments, which often include sensor noise, annotation inaccuracies, and diverse driving conditions—factors that are typically absent in controlled simulation settings.}
Therefore, narrowing the domain gap is a crucial indicator of the synthetic datasets' realism and generalizability~\cite{hu2023simulation}. The primary goal is to assess the realism of SV2V-RSim when exposed to real-world data variations. The experiment consists of the following two parts:

\textbf{Generalization to Real-World Data.} The results of the sim2real experiments are shown in \nzk{Table} \ref{tab:sim2real}. 
We compare the performance of our SV2V-RSim datasets with that of other simulator dataset (OPV2V~\cite{xu2022opv2v}), using baseline methods that are trained solely on different synthetic data and tested on the same real data (V2XReal~\cite{xiang2024v2xreal}). Since OPV2V only provides annotation labels for the car, we conduct detection solely on this category. \nzk{The results of the LiDAR detection reflect geometric authenticity of our assets. For fair comparison, we randomly subsample SV2V-RSim's training data to match OPV2V's volume.} Better detection results indicate a smaller domain gap between the synthetic and real datasets, thereby enhancing their real-world applicability.

Furthermore, to validate the effectiveness of simulation data on real-world data, we pre-train the model on part of SV2V-RSim and then conduct full training on V2XReal. 
The obtained detection results show further improvement compared to training solely on real-world data, demonstrating that high-quality simulation data has a positive impact on the effectiveness of real-world data.
The results of this experiments demonstrate that SV2V-RSim performs better on Sim2Real expriments when faced with real-world data. Without a domain adaptation module, we improve the detection results on real data by enhancing the realism of the synthetic dataset.

\begin{table}[t]
\small
    \centering
    \begin{tabular}{l l c c}
        \toprule
        Train Data & Model      & mAP@0.3 & mAP@0.5 \\
        \midrule
        \multirow{2}{*}{V2XReal} & Where2comm & 53.2   & 51.1    \\
                                    & CoBEVT     & 56.4    & 53.6    \\
        \midrule
        \multirow{2}{*}{OPV2V}     & Where2comm & 10.7    & 8.6    \\
                                    & CoBEVT     & 25.9    & 23.4    \\
        \midrule
        \multirow{2}{*}{Ours} & Where2comm & 32.2    & 27.2    \\
                                    & CoBEVT     & 39.9    & 34.3    \\
        \midrule
         \multirow{2}{*}{V2XReal*} & Where2comm & 54.7    & 52.2    \\
                                & CoBEVT     & 57.8    & 54.3    \\
        \bottomrule
    \end{tabular}
    \caption{Car detection results on the V2XReal test set, with models trained on different datasets (V2XReal, OPV2V and our SV2V-RSim). V2XReal* represents pre-training the model using mixture of SV2V-RSim and V2XReal data and then training it on V2XReal.}
    \label{tab:sim2real}
\end{table}

\textbf{Synthetic Image Quality Assessment}. We also employ no-reference image quality assessment (\textit{i.e.}, MUSIQ~\cite{musiq}, MANIQA~\cite{maniqa}, NIQE~\cite{niqe}, ILNIQE~\cite{ilniqe}) to evaluate the quality of SV2V-RSim by comparing its images with those of OPV2V. 
Similarly, the number of comparisons between the two methods is kept consistent. 
Table~\ref{tab:iqa} presents the results of image quality assessment (IQA) metrics. 
\nzk{As shown in Table~\ref{tab:iqa}, SV2V-RSim consistently achieves higher scores across all NR-IQA metrics, confirming its superiority in visual realism and perceptual quality.}

\setlength{\tabcolsep}{3.5pt}
\begin{table}[t]
\small
    \centering
    \begin{tabular}{lcccc}
        \toprule
        Method & MUSIQ\textuparrow & MANIQA\textuparrow & NIQE\textdownarrow & ILNIQE\textdownarrow \\
        \midrule
        OPV2V       & 58.61 & 0.33 & 3.61 & 24.63 \\
        SV2V-RSim   & \textbf{62.68} & \textbf{0.36} & \textbf{3.10} & \textbf{23.37} \\
         ~ &+7.0\% &+9.1\% &-14.1\% & -5.12\% \\
        \bottomrule
    \end{tabular}
    \caption{Comparison of image quality metrics between OPV2V and SV2V-RSim. Arrows indicate whether higher or lower is better.}
    \label{tab:iqa}
\end{table}

\section{Benchmark and Experiments}
We conduct systematic experiments on collaborative perception in autonomous driving, including both independent and collaborative 3D object detection, as well as agent selection. In these experiments, various evaluation metrics are utilized to assess the model's performance across different experiments. The benchmarks for each experiment will be presented in detail in the following subsections.



\subsection{Experiment Configurations}
The experiments are conducted on an NVIDIA 4090 GPU. The dataset is split into a 7:3 ratio for training and testing. During training, we set the number of epochs to 30, using Adam optimizer with an initial learning rate of 0.0002. All benchmark experiment results are averaged over three runs.

\subsection{Proposed Baseline: Select Vehicles Adaptively}

The high density of SV2V-RSim, averaging 16.2 collaborative agents per scenario, renders traditional broadcast-based feature sharing computationally prohibitive and inefficient in terms of communication bandwidth. To address this limitation, we introduce the Select Vehicles Adaptively (SVA) module as a baseline. SVA operates via a two-stage communication protocol designed to acquire spatial features based on actual perceptual demand. Rather than indiscriminately aggregating data from all nearby agents, SVA filters collaborative features by aligning the observation capabilities of surrounding vehicles with the ego vehicle's spatial blind spots, thereby improving the trade-off between detection accuracy and bandwidth consumption.


\subsubsection{Spatial Confidence Estimation}
 Initially, each vehicle $v_j$ encodes its raw point cloud $\mathcal{L}_j$ into a BEV feature $f_j = \Phi_{enc}(\mathcal{L}_j) \in \mathbb{R}^{C \times H \times W}$ using PointPillars. To minimize initial communication overhead, $v_j$ predicts a lightweight spatial confidence map $m_j \in [0, 1]^{H \times W}$, representing regions with reliable observations. This compact map, alongside the vehicle's 6-DoF pose $P_j$, is transmitted to the ego vehicle as a preliminary metadata packet $\mathcal{D}_{j \rightarrow ego} = \{m_j, P_j\}$.



\subsubsection{Adaptive Feature Selection}
Upon receiving the metadata, the ego vehicle evaluates the necessity of requesting dense features. We initialize a query vector $q$, where the ego vehicle's features act as a unified whole when attending to other agents: $q = \text{attn}(q_{init}, f_{ego})$. SVA then calculates a spatial selection mask $M_{i \rightarrow ego} \in [0, 1]^{H \times W}$ for each candidate agent $i$ by aligning the ego's query with the collaborator's metadata:

\begin{equation}
    M_{i \rightarrow ego} = \sigma(\text{MLP}(\text{Concat}(q, \tilde{m}_{i \rightarrow ego}, \text{Embed}(P_i))))
\end{equation}
where $\tilde{m}_{i \rightarrow ego}$ denotes the confidence map of vehicle $i$ warped to the ego coordinate system, $\sigma(\cdot)$ is the sigmoid function, and $\text{Embed}(\cdot)$ extracts the relative positional features. This mechanism physically ensures that transmission weights are activated primarily in regions where the ego vehicle lacks visibility (its own blind spots) but the collaborator provides high-confidence data.

\subsubsection{Point-wise Feature Integration}
To further compress bandwidth, SVA performs point-wise data transmission across multiple vehicles rather than conventional binary agent-level selection. Specifically, the ego vehicle broadcasts the coordinates corresponding to the non-zero regions of the calculated weight map $M_{i \rightarrow ego}$ back to the respective collaborators. The collaborators then filter their dense features to transmit only these queried spatial regions, yielding the masked feature $\tilde{f}_{i \rightarrow ego} = \text{Warp}(f_i) \odot M_{i \rightarrow ego}$. Finally, the ego vehicle aggregates these tailored features:
\begin{equation}
  \mathcal{F}_{fuse} = \Phi_{fusion}(\mathcal{F}_{ego}, \{\tilde{f}_{i \rightarrow ego}\}_{i=1}^N)
\end{equation}
where $N$ is the number of collaborative vehicles. The final 3D bounding box predictions are generated via the detection head $\mathcal{Y}' = \Phi_{dec}(\mathcal{F}_{fuse})$. By strictly filtering out features that fall outside the non-zero regions of the weight map, SVA fundamentally reduces bandwidth consumption while maximizing perceptual gain in heavily occluded scenes.

\subsection{Benchmark Comparison}
\begin{table*}[t]
\small
    \centering
    \begin{tabular}{l cc cc cc cc cc}
        \toprule
        Models & \multicolumn{2}{c}{AP$_{M}$@IoU} & \multicolumn{2}{c}{AP$_{N}$@IoU} & \multicolumn{2}{c}{AP$_{P}$@IoU} & \multicolumn{2}{c}{AP$_{S}$@IoU} & \multicolumn{2}{c}
        {Bandwidth}\\
        & 0.3 & 0.5 & 0.3 & 0.5 & 0.3 & 0.5 & 0.3 & 0.5\\
        \midrule
        No Fusion & 32.6 & 28.7 & 19.2 & 12.7 & 3.3 & 0.6 & 5.7 & 3.2 & 0\\
        Late Fusion & 48.1 & 42.6 & 25.2 & 13.5 & 4.2 & 1.1 & 10.4 & 7.0 & 12.91\\
        Early Fusion & 57.9 & 56.0 & 30.9 & 23.7 & 4.9 & 1.5 & 13.9 & 8.1 & 25.54\\
        \midrule
        F-Cooper & 56.7 & 53.8 & 45.9 & 38.7 & 15.4 & 4.7 & 32.1 & 21.2 & 27.85\\
        CoBEVT & 57.9 & 55.6 & 52.4 & 44.3 & 23.3 & 12.7 & 43.2 & 30.2 & 28.85\\
        Where2comm & 62.9 & 60.7 & 37.3 & 25.6 & 13.7 & 3.7 & 25.3 & 12.0 & 26.72\\ 
        SVA(ours) & 65.2 & 62.6 & 49.5 & 38.2 & 21.8 & 11.3 & 41.3 & 27.6 & 25.53 \\
        \bottomrule
    \end{tabular}
    \caption{Benchmark results of recent cooperative perception methods and our proposed method. Our method achieves consistently superior performance-bandwidth trade-off. The four major categories are evaluated independently for detection accuracy. The average communication bandwidth is measured in $log_2$ scale. AP$_{M}$, AP$_{N}$, AP$_{P}$ and AP$_{S}$ represent the detection results for motor vehicles, non-motor vehicles, pedestrians, and static objects, respectively.}
    \label{tab:benchmark_results}
\end{table*}
We evaluate the performance of various fusion strategies and recent models for multi-class 3D object detection, focusing on detection accuracy and efficiency in complex scenarios. The dataset includes four categories: motor vehicles, non-motor vehicles, person, and static objects, with Average Precision and bandwidth consumption as core metrics. The results are shown in Table~\ref{tab:benchmark_results}.

The single-sensor baseline demonstrates limited overall performance, particularly in motor vehicle detection. In contrast, fusion-based methods exhibit significant improvements. Late Fusion and early fusion enhance detection accuracy by integrating features at different stages. Late fusion requires less communication bandwidth, while early fusion achieves better performance. Among advanced models, F-Cooper, CoBEVT, and Where2comm attains the higher accuracy. However, all methods struggle with the person category, indicating a limited generalization to smaller or irregularly shaped objects. Where2comm highlights a trade-off between communication efficiency and multi-class robustness. The transmission amount is greatly reduced while the prediction accuracy is guaranteed. However, due to the imperfections in the selection strategy and feature mask processing, Where2comm experiences a decline in detection performance for small objects. Since F-Cooper and CoBEVT do not perform selection, their bandwidth consumption is significantly high.

The proposed SVA model achieves outstanding results in motor vehicle detection, while also excelling in static object detection. Compared to Where2Comm, SVA performs better in terms of communication bandwidth and feature processing. Although its performance on non-motor vehicles lags behind CoBEVT, SVA maintains competitive overall accuracy without additional bandwidth overhead, demonstrating effective multitarget perception under resource constraints. 

\subsubsection{Advantage of Active Spatial Querying}
\begin{figure}[t]
  \centering
  \includegraphics[width=3.2in]{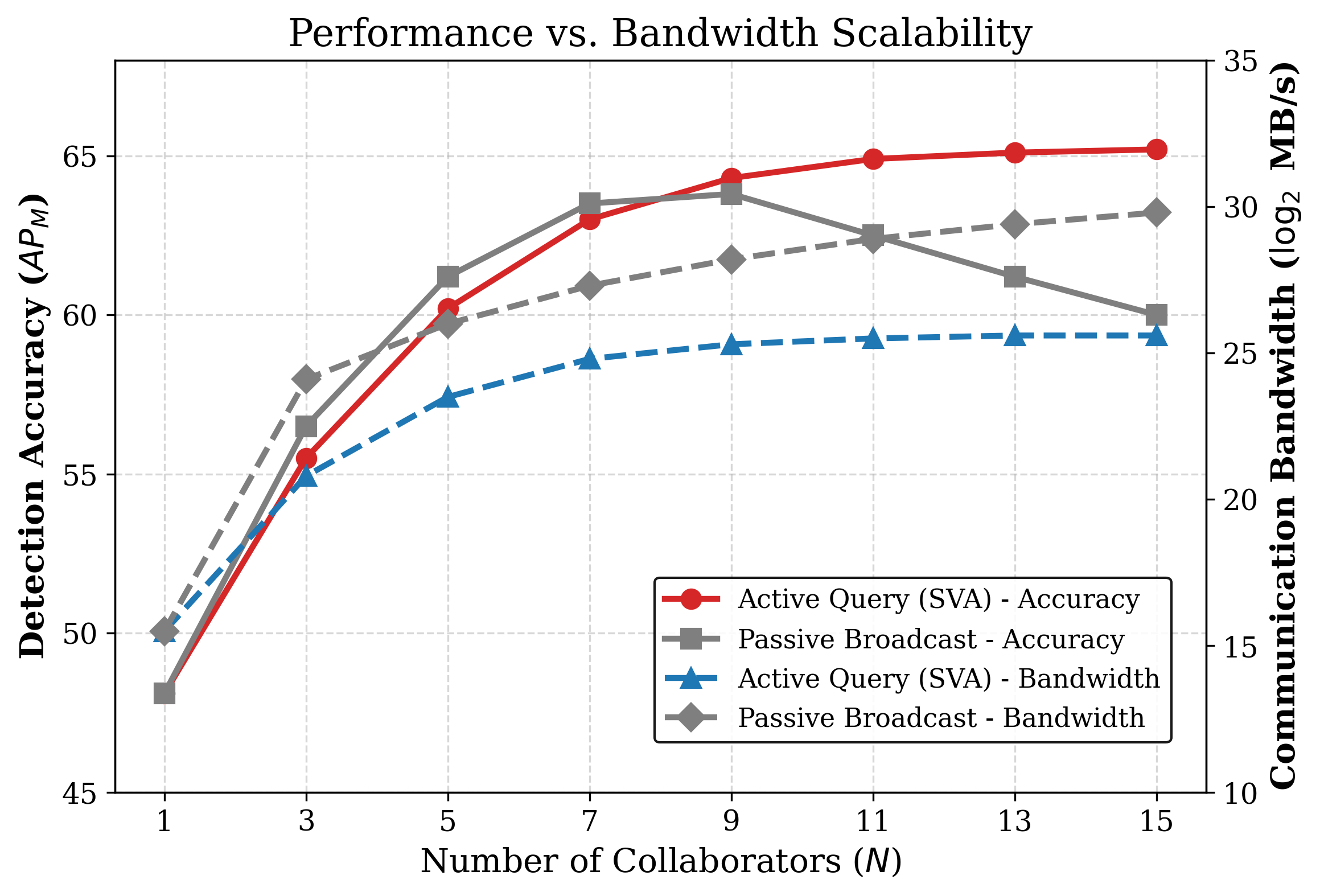}
   \caption{Performance and bandwidth scalability.}
   \label{fig:scalability_curve}
\end{figure}

As shown in Figure \ref{fig:scalability_curve}, passive broadcasting in dense scenarios causes linear bandwidth explosion and accuracy degradation from overlapping spatial noise. Conversely, SVA actively queries only ego blind spots, efficiently capping bandwidth while maximizing accuracy.

\section{Conclusion}
\label{sec:conclusion}
In this paper, we introduce SV2V-RSim, a large-scale, multi-modal, near-realistic simulation dataset designed to bridge the gap between simulated and real-world cooperative perception scenarios. To validate the effectiveness of our dataset, we conduct sim2real experiments. The results demonstrate that training with SV2V-RSim improves accuracy by 16.25 compared to OPV2V when testing on real-world data. Moreover, pre-training the model on part of SV2V-RSim and fully training on real data further enhances detection performance. Additionally, we propose the Select Vehicles Adaptively (SVA) module within our benchmark framework for cooperative multi-class 3D object detection. This approach effectively balances detection accuracy and communication bandwidth in collaborative perception systems. Our method outperforms existing techniques, improving detection accuracy while reducing communication bandwidth.

\clearpage  


%
%
\bibliographystyle{splncs04}
\bibliography{main}
\end{document}


\title{SV2V-RSim: A Comprehensive Benchmark for Self-Selective V2V Cooperative
Perception with Near-Realistic Data} 

\titlerunning{Abbreviated paper title}

\author{First Author\inst{1}\orcidlink{0000-1111-2222-3333} \and
Second Author\inst{2,3}\orcidlink{1111-2222-3333-4444} \and
Third Author\inst{3}\orcidlink{2222--3333-4444-5555}}

\authorrunning{F.~Author et al.}

\institute{Princeton University, Princeton NJ 08544, USA \and
Springer Heidelberg, Tiergartenstr.~17, 69121 Heidelberg, Germany
\email{lncs@springer.com}\\
\url{http://www.springer.com/gp/computer-science/lncs} \and
ABC Institute, Rupert-Karls-University Heidelberg, Heidelberg, Germany\\
\email{\{abc,lncs\}@uni-heidelberg.de}}

\maketitle

\section{Unreal Engine 5-based Simulation}
\label{sec:simulation}
In this section, we provide more details about our Unreal Engine 5-based Simulation, including the 3D assets, maps, traffic flows, weather and sky systems.

\subsection{3D assets}
\label{sec:supp_model}
\begin{figure*}[t]
    \centering
    \begin{minipage}{0.45\textwidth}
        \centering
        \includegraphics[width=\linewidth]{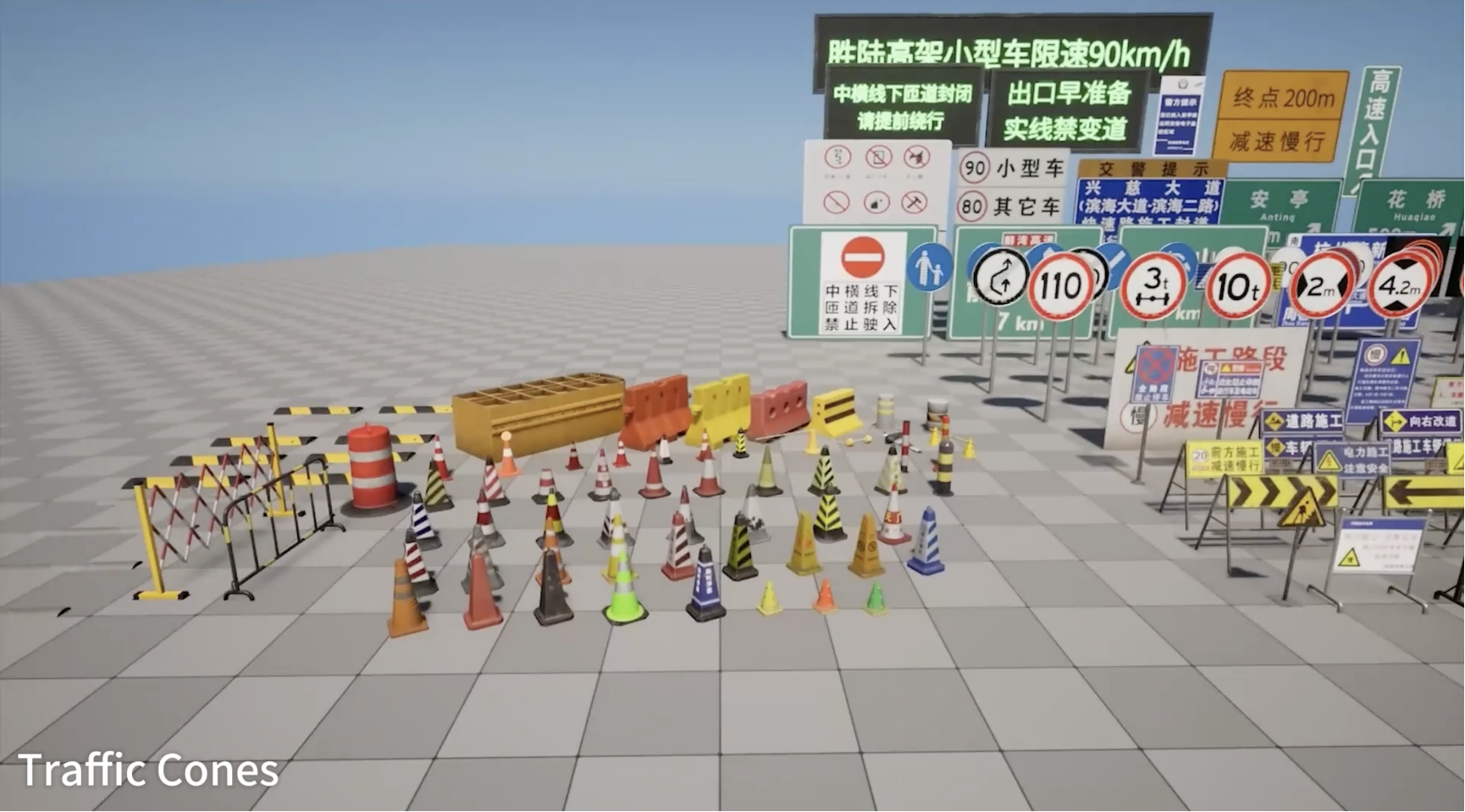}
    \end{minipage}\hspace{0.1cm}  
    \begin{minipage}{0.45\textwidth}
        \centering
        \includegraphics[width=\linewidth]{images/traffic_signs(e).png}
    \end{minipage}

    \vspace{0.1cm}

    \begin{minipage}{0.45\textwidth}
        \centering
        \includegraphics[width=\linewidth]{images/traffic_signs(c).png}
    \end{minipage}\hspace{0.1cm}  
    \begin{minipage}{0.45\textwidth}
        \centering
        \includegraphics[width=\linewidth]{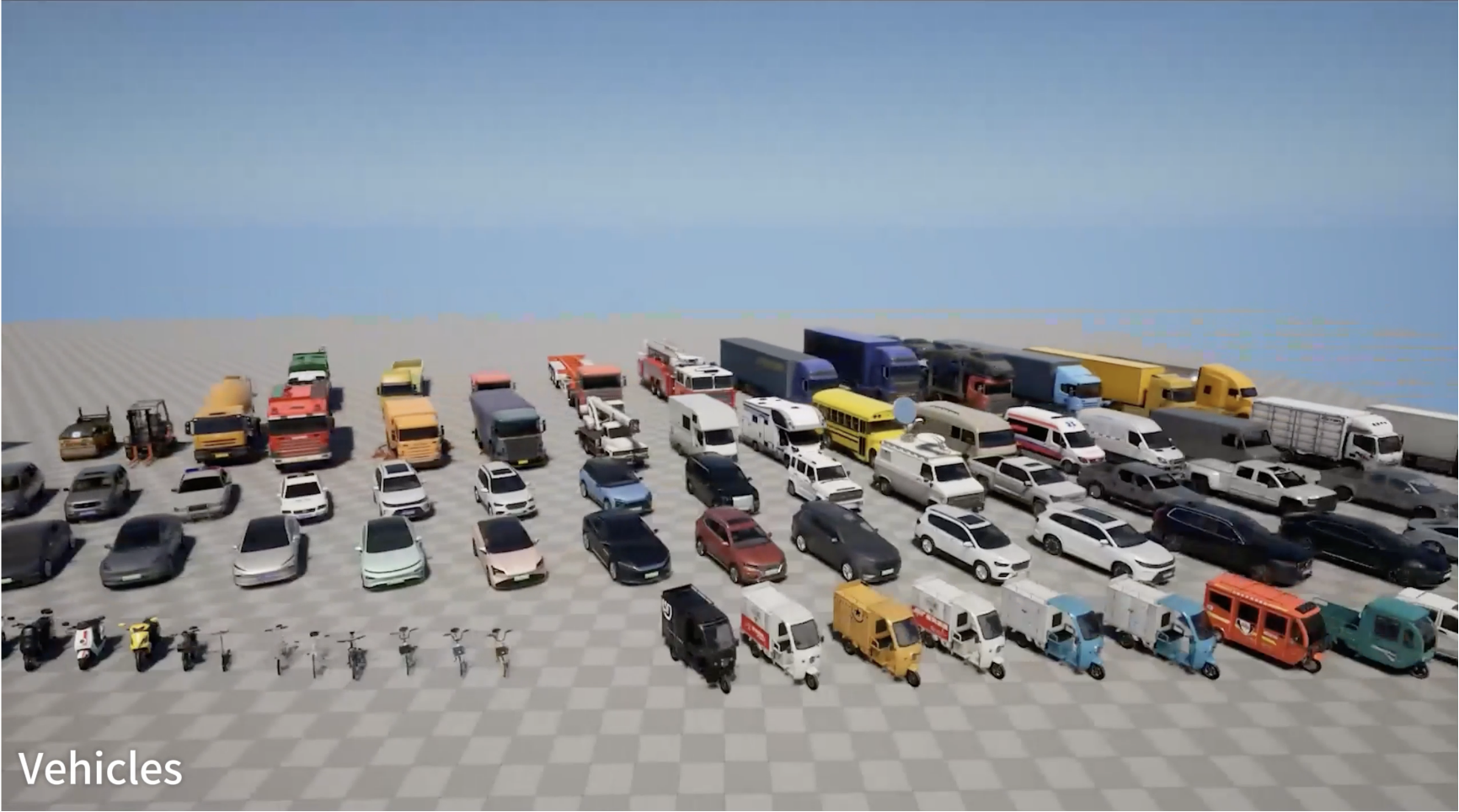}
    \end{minipage}
    \caption{Part of our autonomous driving asset library.}
    \label{fig:assets}
\end{figure*}
We use high-quality 3D assets with diverse surface materials and accurate 3D structures, which make our dataset more realistic in images and more accurate in point clouds. The utility of the more realistic dataset has already been demonstrated in the experiments. Figure~\ref{fig:assets} shows part of our autonomous diriving asset library, including traffic cones, traffic signs and dynamic objects. Table~\ref{tab:assets} summarizes the assets employed. 

\begin{table}[b]
    \centering
    \begin{tabular}{lcl}
        \toprule
        \textbf{Category} & \textbf{Assets} & \textbf{Sub Category} \\ \midrule
        Motor & 91 & Car, Trailer, Truck, Van\\
        & &Pickup, SpecialShaped, Bus\\ 
        & &Engineering, MediumSized \\
        Non-Motor & 9 & Bicycle, Tricycle, Ebike \\
        Static Objects & 10 & WarningPole, Obstacle \\
        Person & 17 & Pedestrian, E-biker\\
        & &Motorcyclist \\
        \bottomrule
    \end{tabular}
    \caption{Summary of asset categories used, listing the total \textit{Number} of each based asset model \textit{Category} along with specific \textit{Sub Category} types.}
    \label{tab:assets}
\end{table}

\subsection{Maps}
\label{sec:supp_map}
\begin{figure*}[!htbp]
    \centering
    \begin{minipage}{0.45\textwidth}
        \centering
        \includegraphics[width=\linewidth]{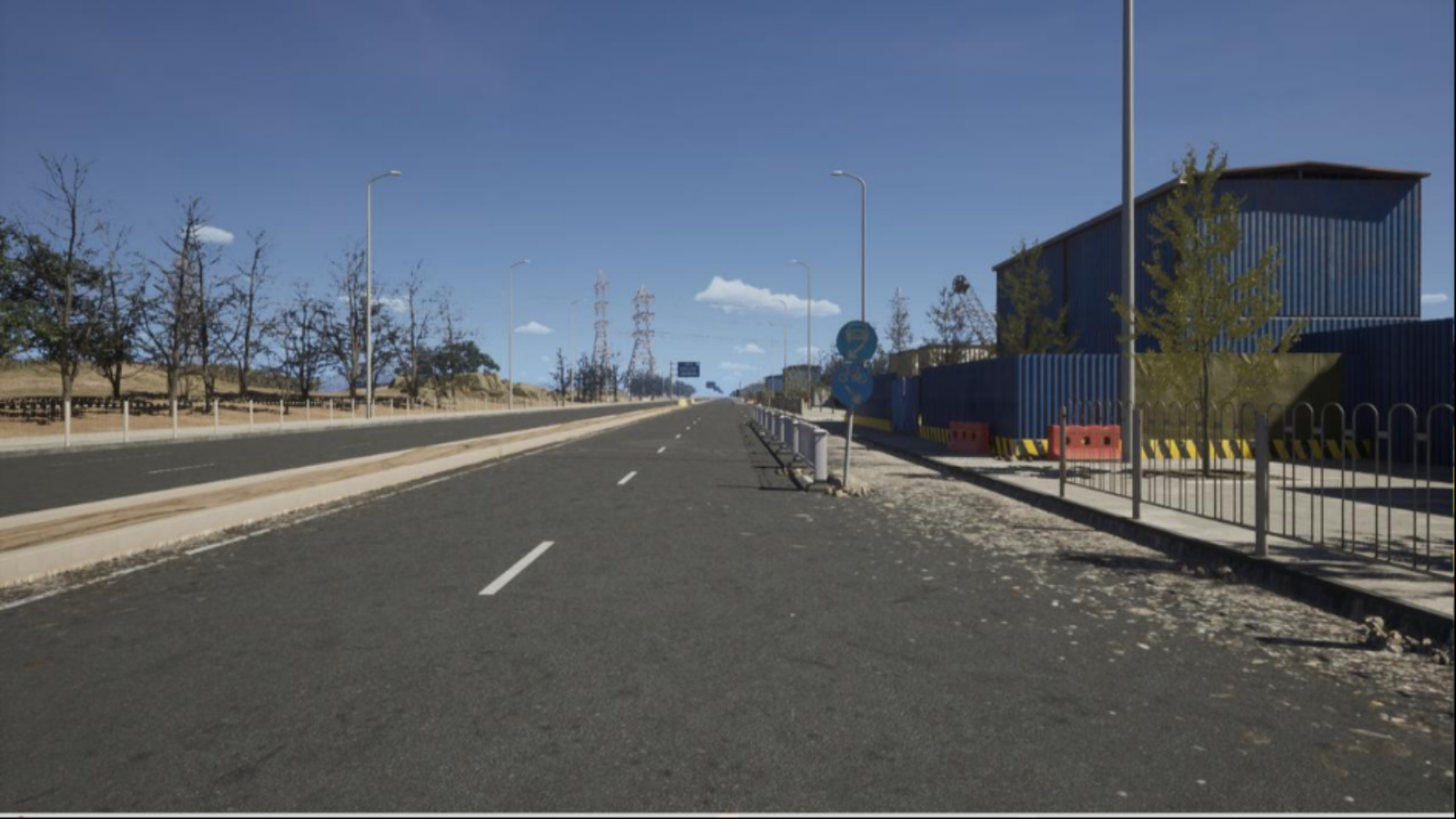}
        \subcaption{Suburban}\label{suburban}
    \end{minipage}\hspace{0.1cm}
    \begin{minipage}{0.45\textwidth}
        \centering
        \includegraphics[width=\linewidth]{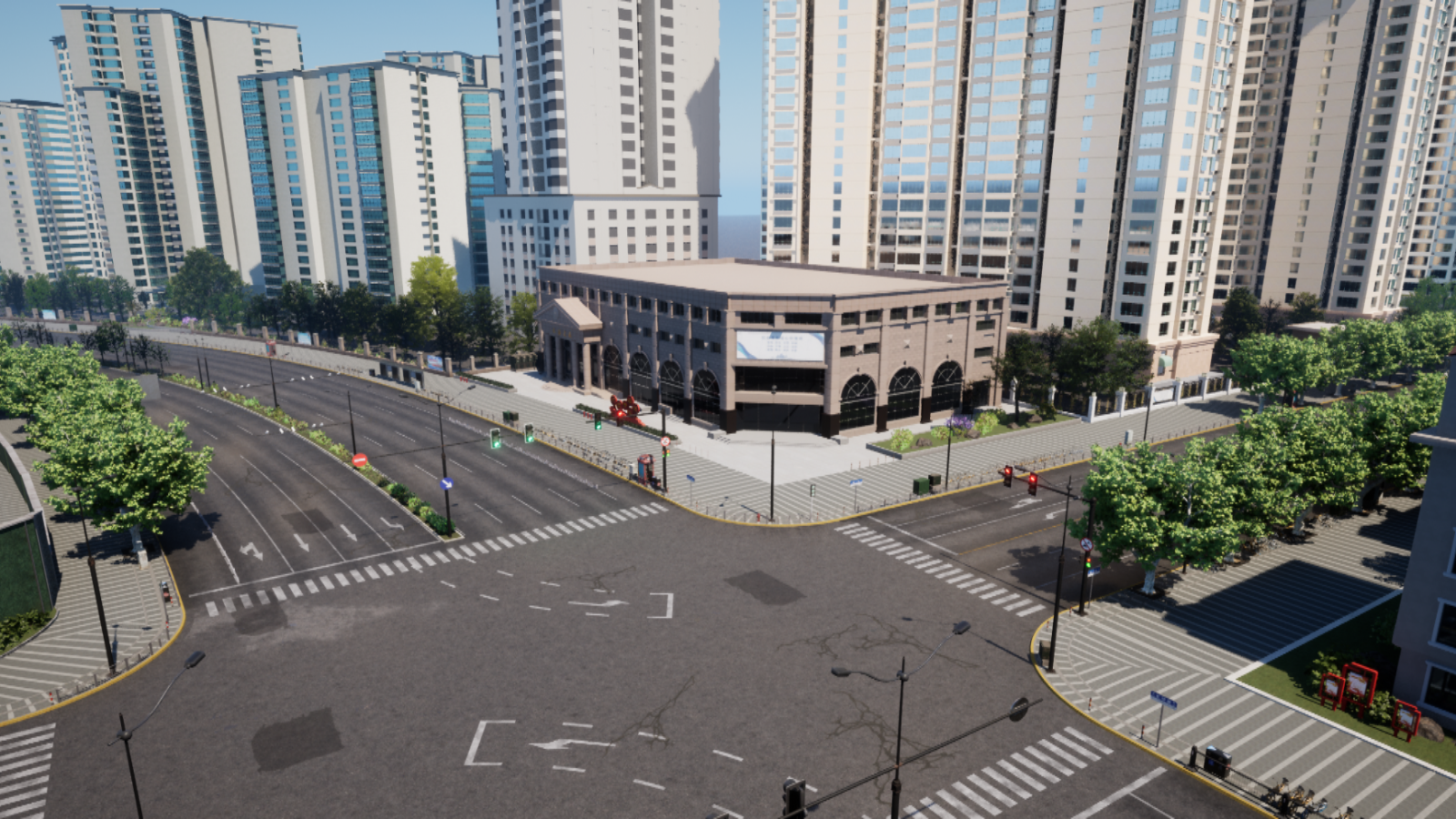}
        \subcaption{Urban}\label{urban}
    \end{minipage}

    \vspace{0.1cm}

    \begin{minipage}{0.45\textwidth}
        \centering
        \includegraphics[width=\linewidth]{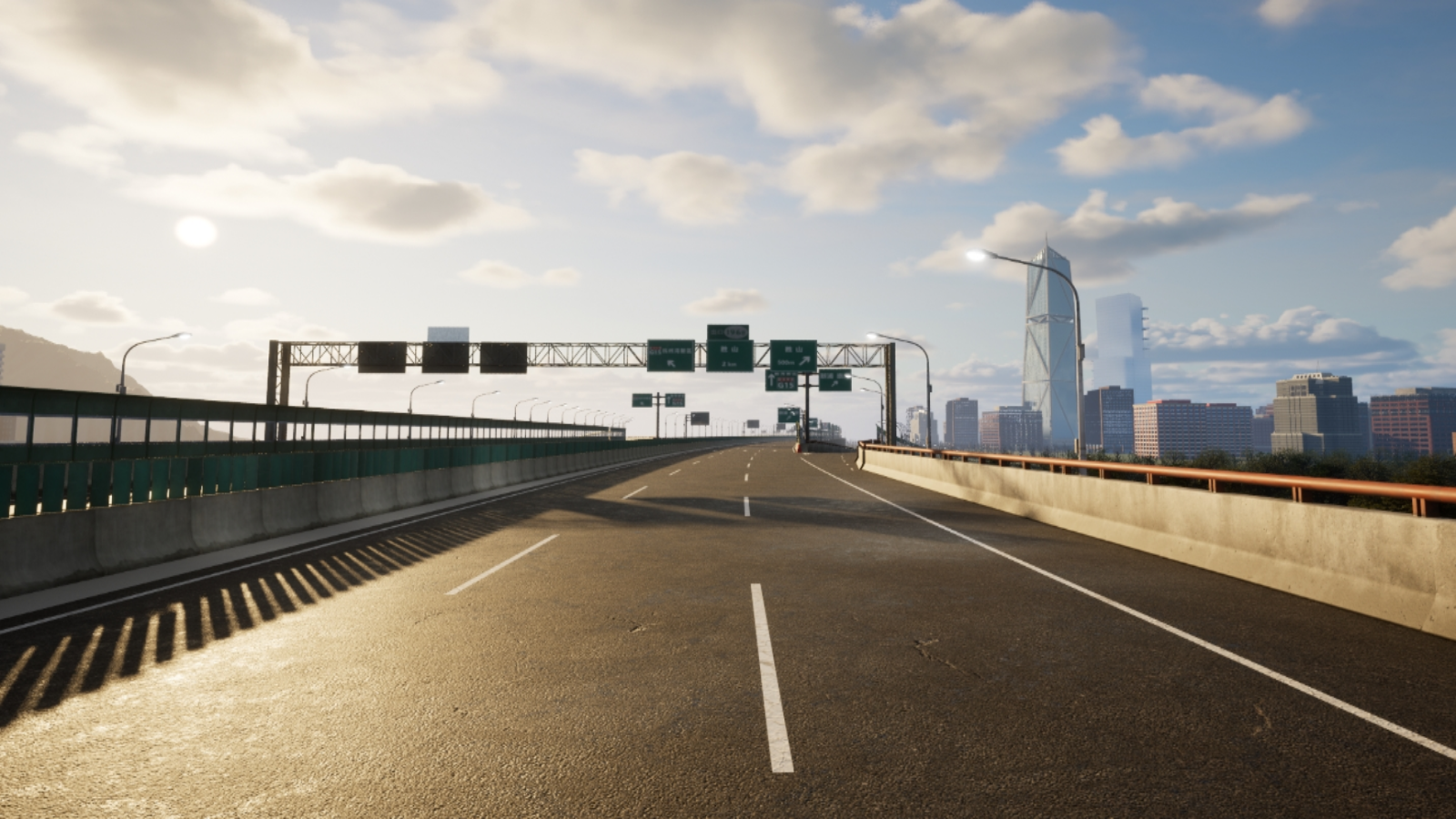}
        \subcaption{Highway}\label{highway}
    \end{minipage}\hspace{0.1cm}
    \begin{minipage}{0.45\textwidth}
        \centering
        \includegraphics[width=\linewidth]{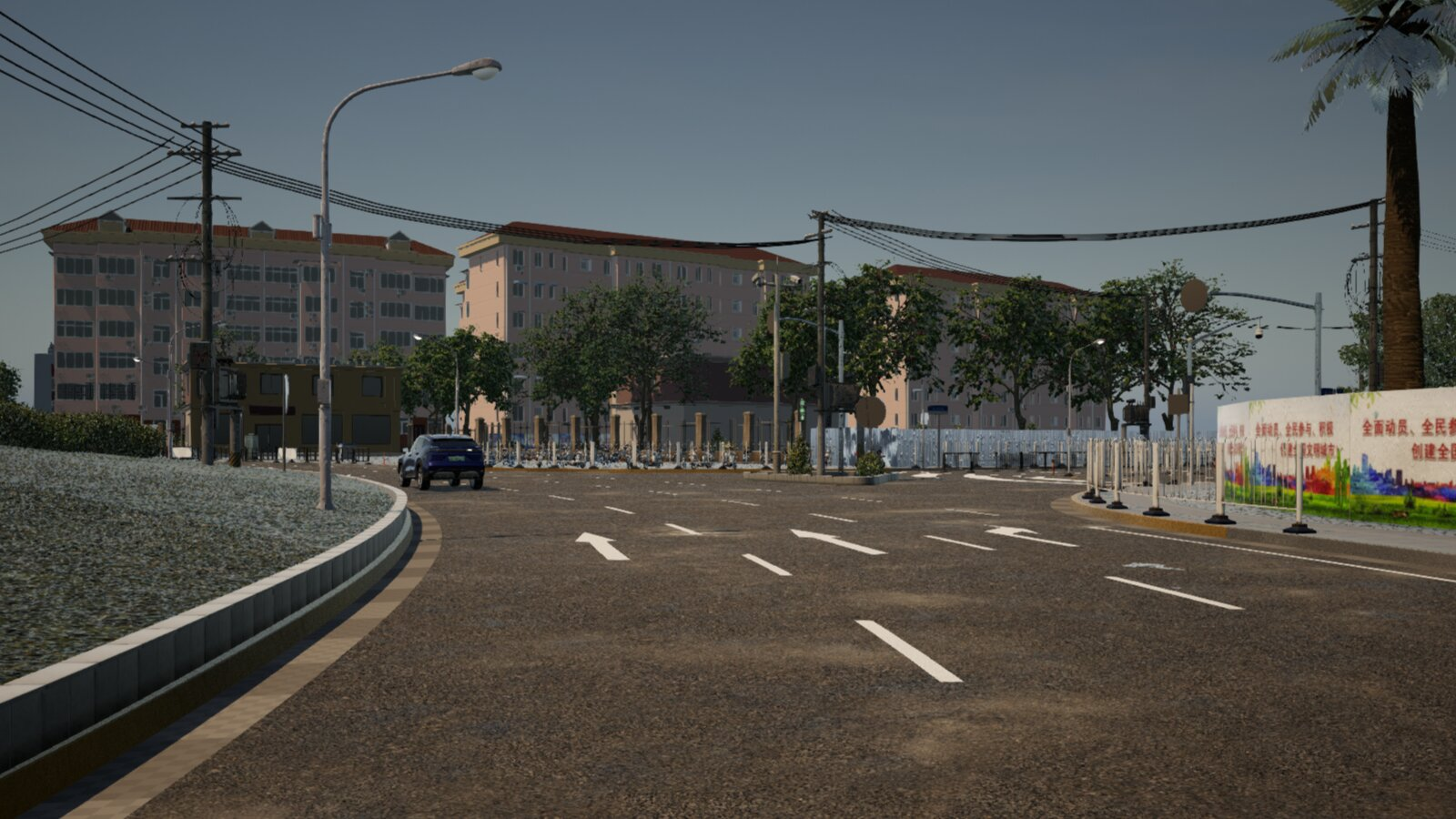}
        \subcaption{Roundabout}\label{roundabout}
    \end{minipage}
    \caption{Rendering visualization of our used maps.}
    \label{fig:maps}
\end{figure*}

Our proposed dataset uses four maps, where Figure \ref{fig:maps} displays each scene, including urban areas, suburban regions and highways. Each map is specially designed to closely resemble real-world scenarios. In urban areas, we have placed multiple shared bicycles on the sidewalks; in suburban areas, vehicles are randomly parked along the roadside; and on the highway, we have placed lots of corresponding traffic signs. What's more, each map contains a variety of differentiated driving segments. For example, the highway is configured with on- and off-ramp segments.

\subsection{Traffic flows} 
\label{sec:traffic_flow}
To mirror the intricacies of real-world traffic conditions, we refined automatically generated traffic flows with detailed custom adjustments. For each map, multiple scene files are generated along every driving route to capture a wide range of scenarios. Moreover, along the same route, we simulate an extensive array of driving behaviors for both the ego vehicle and other vehicles—including acceleration, deceleration, lane changing, overtaking, and turning. Notably, we design scenarios with dynamic collaborators by introducing variations in speed and lane occupancy. For instance, an overtaking maneuver or another vehicle departing from the current lane results in minor changes, whereas the ego vehicle changing lanes leads to major shifts in the collaborative dynamics.

\subsection{Weather and Sky Systems}
\label{sec:supp_weather}
\begin{figure*}[!htbp]
\centering
    \centering{\includegraphics[width=1\linewidth]{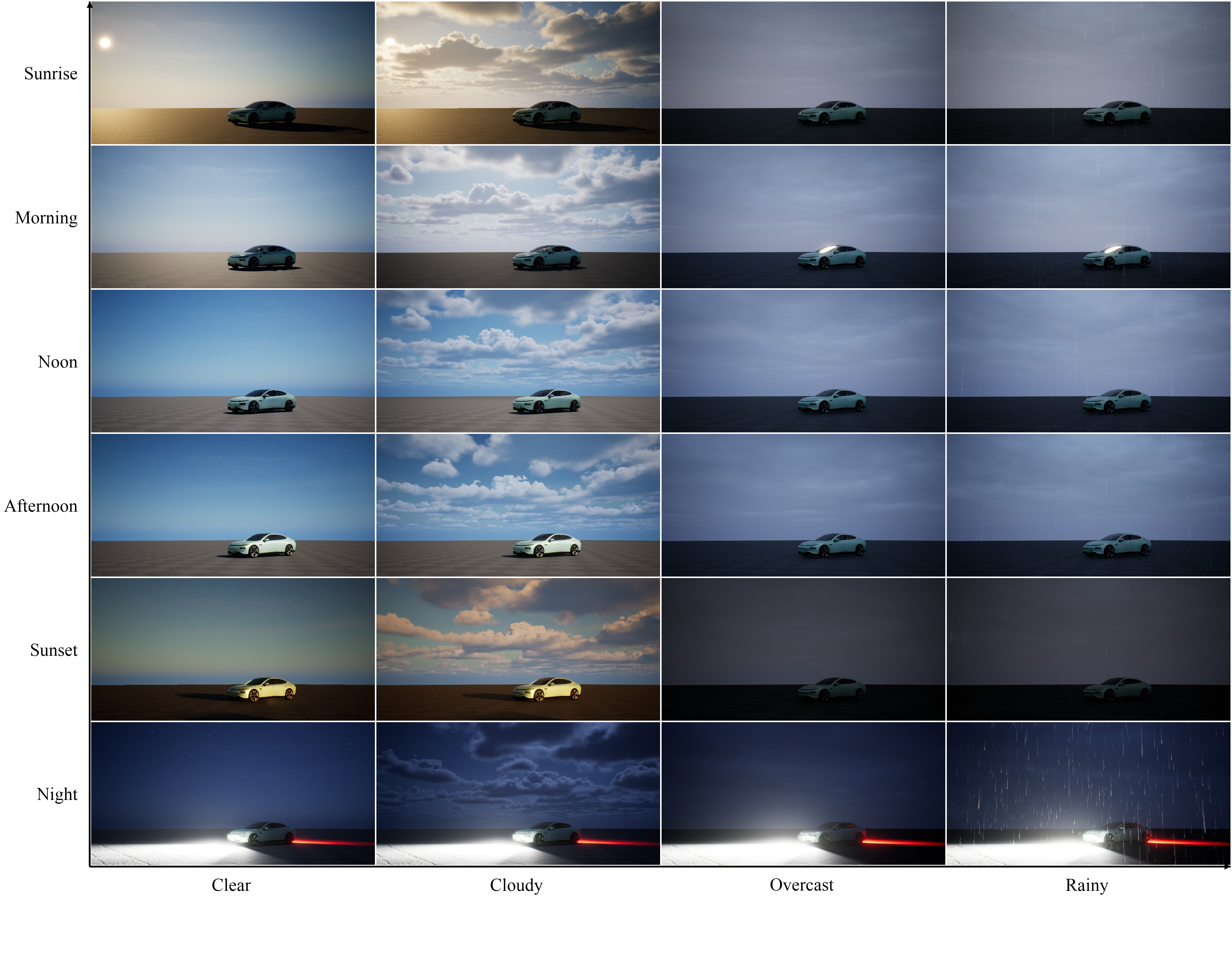}}
\caption{Rendered images of different weathers and skys}
\label{fig:weather_sky}
\end{figure*}

In this section, we provide a detailed demonstration of the weather and sky system through various rendered images. Figure~\ref{fig:weather_sky} shows the rendered images for each combination of weather and sky. Moreover, if it is nighttime, the car lights will be turned on.

\begin{figure}[!htbp]
  \centering
  \includegraphics[width=3.2in]{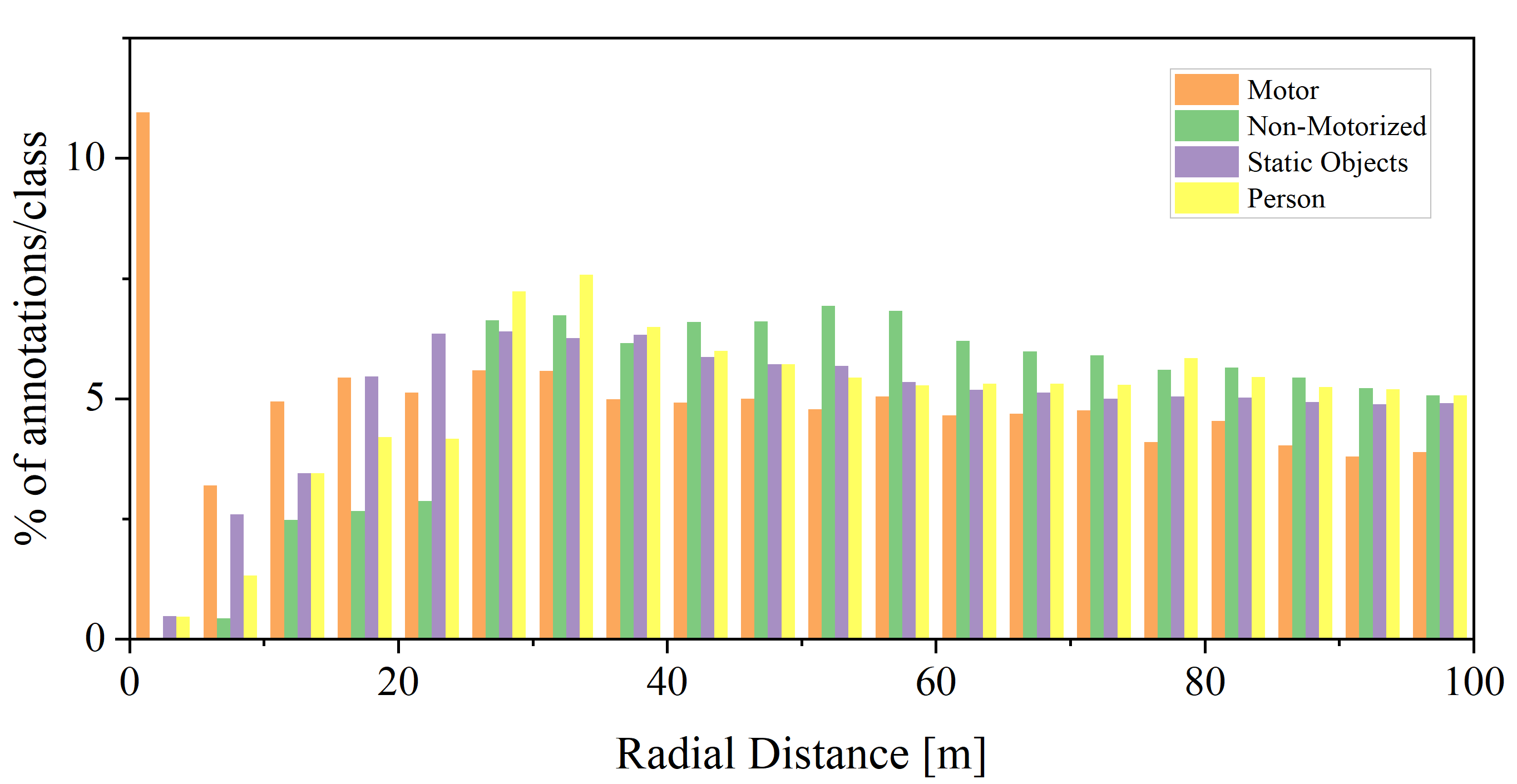}

   \caption{Radial distance of objects for four categories from the ego vehicle.}
   \label{fig:distance}
\end{figure}

\begin{figure}[b]
  \centering
  \includegraphics[width=3.2in]{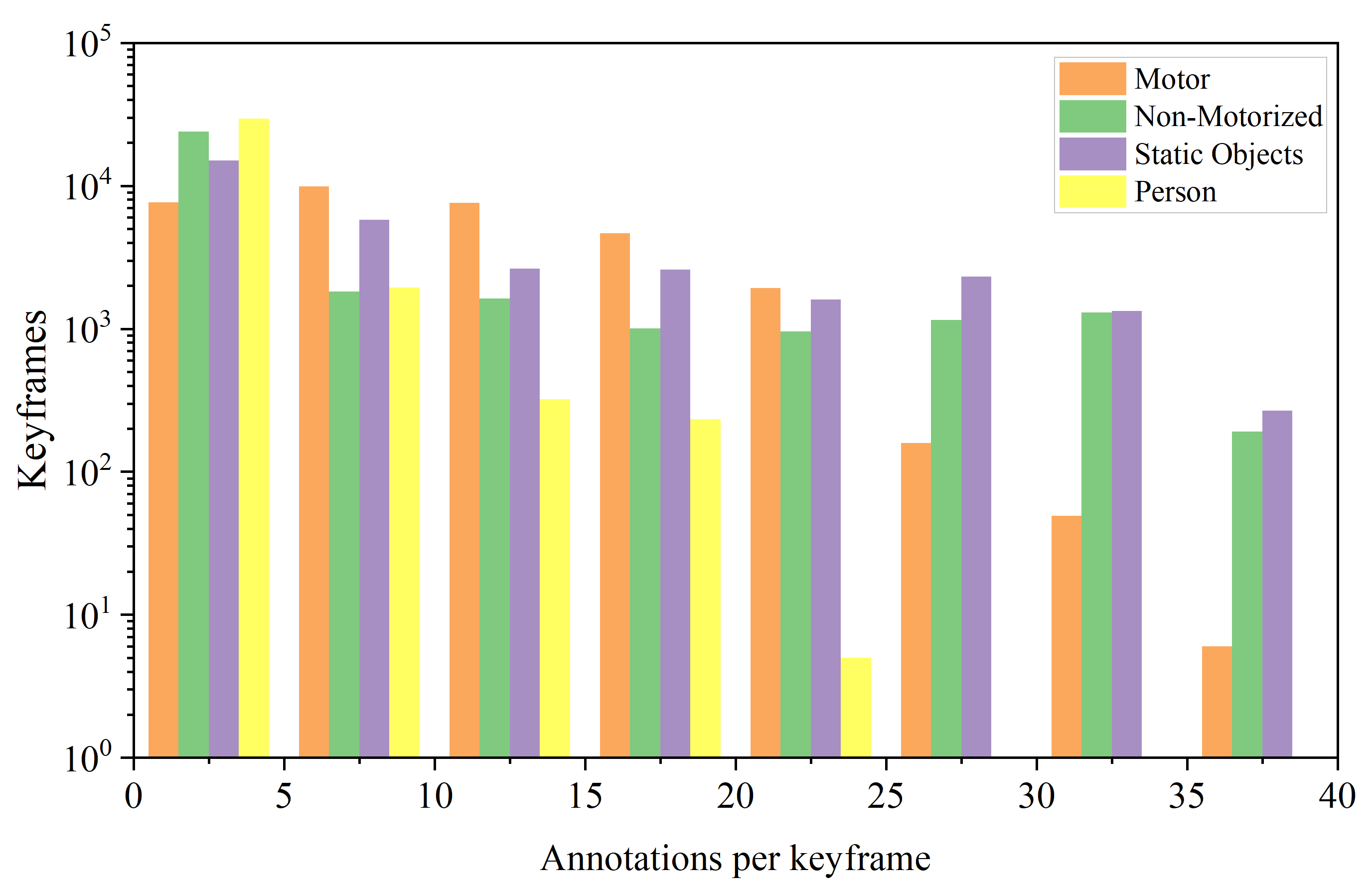}

   \caption{Category count in each keyframe for four categories.}
   \label{fig:keyframe}
\end{figure}

\section{The SV2V-RSim dataset}
In this section, we provide more details about the SV2V-RSim dataset. We first provide our annotation approach and conduct a more detailed data analysis of the dataset. Then, we detail the model configurations and results of the sim-to-real experiments and cooperative perception methods.

\subsection{Data Annotation}
\label{sec:data_annotation}

\subsubsection{Different tasks} We provide data, including RGB images, depth images, segmentation images, LiDAR, and segment LiDAR, for different tasks. The semantic segmentation data can be obtained by querying the corresponding instance types and performing a mapping.

\subsubsection{Annotation JSON} In the annotation JSON, we provide rich object annotation information. 

First, the sensor’s intrinsic parameters are provided, with the rotation represented as a quaternion (in the $wxyz$ order). The extrinsic parameters, expressed in the ego vehicle’s coordinate system, are also given. Additionally, the position and rotation information of the ego vehicle and sensors in the world coordinate system are provided.

For each object label, we annotate its 10-dimensional
cuboid containing x, y, z for the cuboid center, width, length, height for the cuboid extent dimension, and quaternion (in the $xyzw$ order) for the cuboid orientation.

\begin{figure*}[!htbp]
    \centering
    \begin{minipage}{0.24\textwidth}
        \centering
        \includegraphics[width=\linewidth]{images/data_analysis/motor_size.png}
        \subcaption{Motor Vehicle}\label{motor_size}
    \end{minipage}\hspace{0.1cm}
    \begin{minipage}{0.24\textwidth}
        \centering
        \includegraphics[width=\linewidth]{images/data_analysis/non_motorized_size.png}
        \subcaption{Non-motorized Vehicle}\label{non_motorized_size}
    \end{minipage}\hspace{0.1cm}
    \begin{minipage}{0.24\textwidth}
        \centering
        \includegraphics[width=\linewidth]{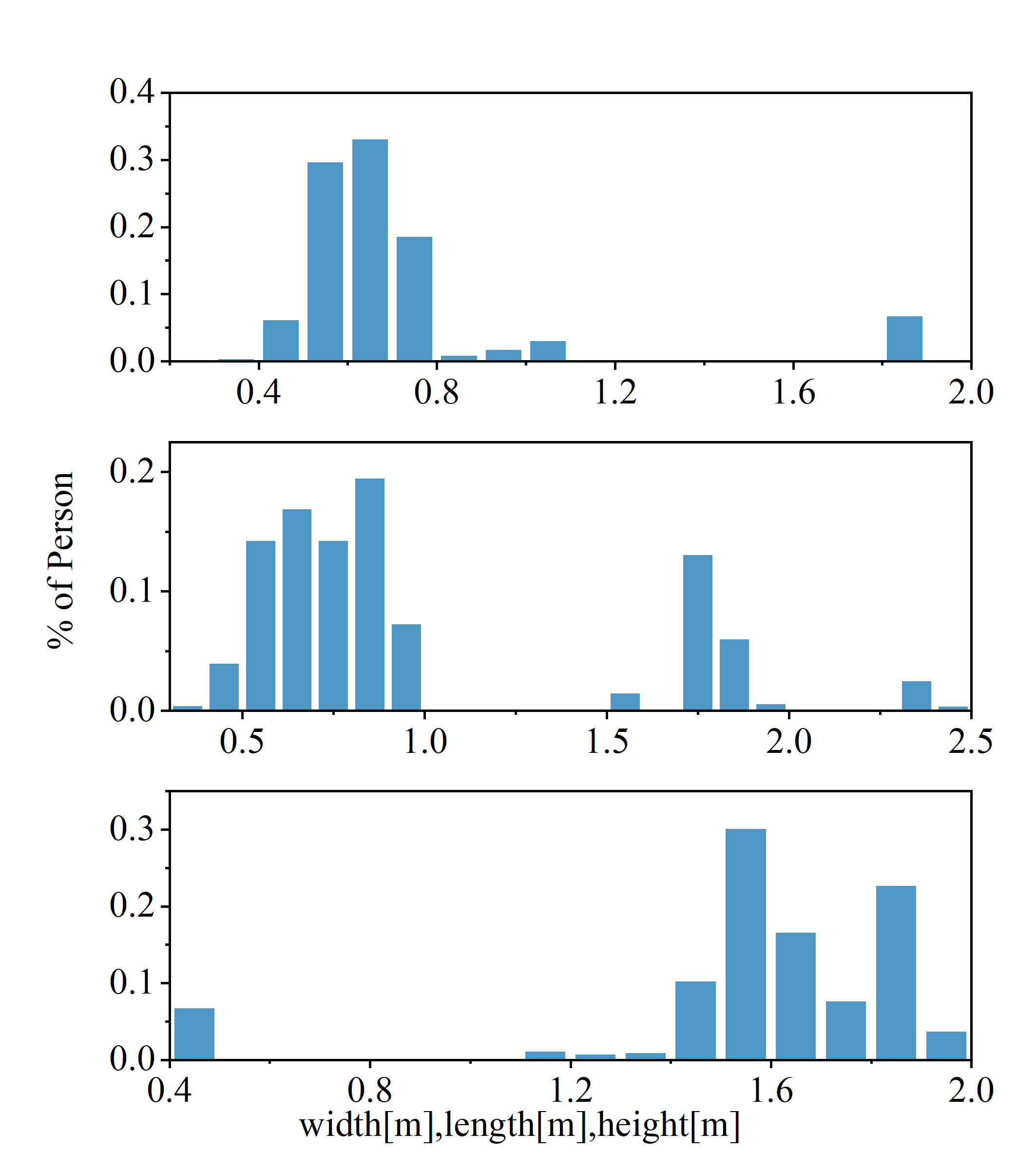}
        \subcaption{Person}\label{person_size}
    \end{minipage}\hspace{0.1cm}
    \begin{minipage}{0.24\textwidth}
        \centering
        \includegraphics[width=\linewidth]{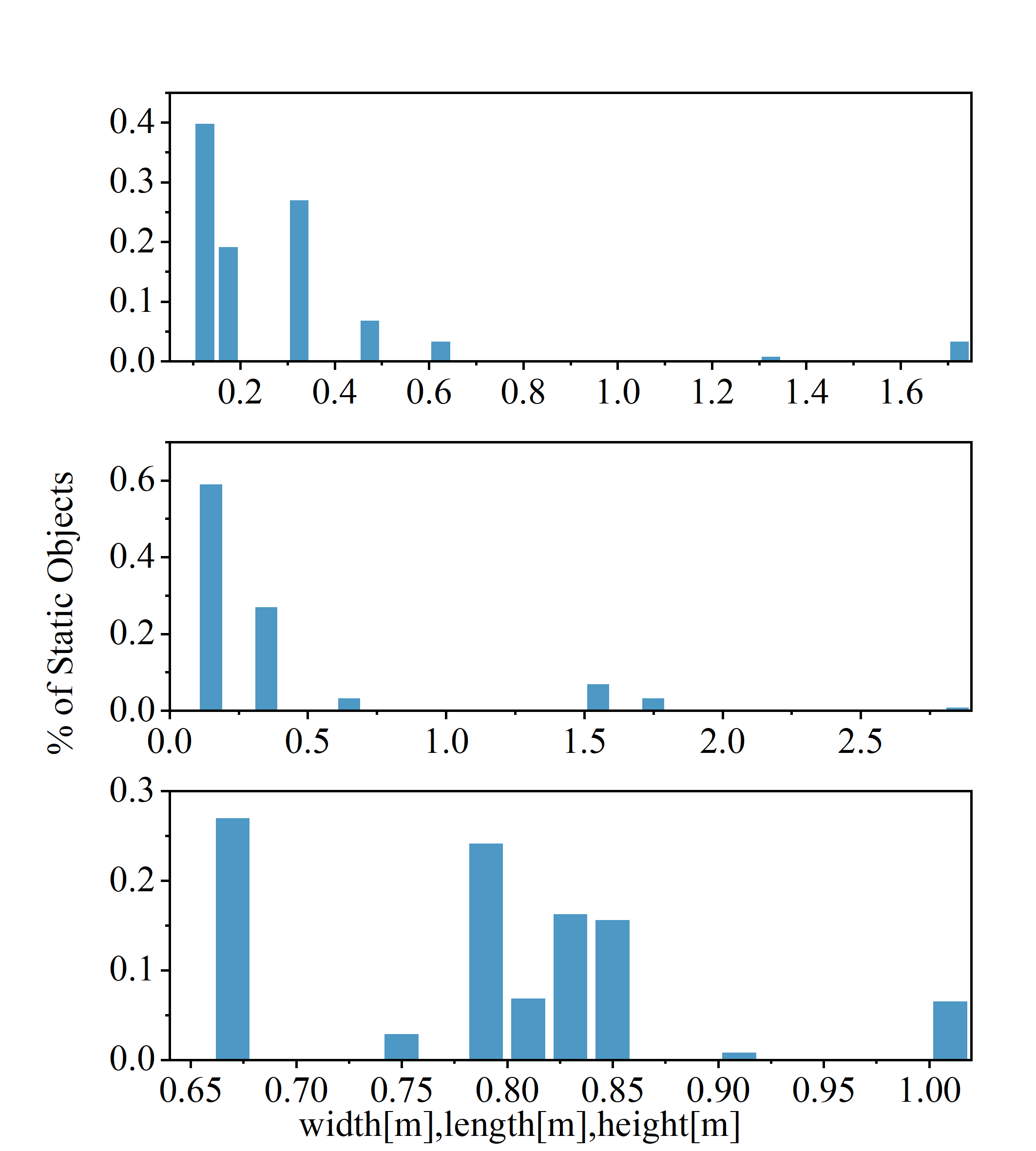}
        \subcaption{Static Object}\label{static_object_size}
    \end{minipage}
    \caption{Bounding box size distributions for four categories.}
    \label{fig:size_dis}
\end{figure*}

\subsection{Data Analysis}
\label{sec:detailed_data_analysis}
Here we provide more analysis about our proposed dataset. Through more detailed data analysis, the data distribution and generalization effects of SV2V-RSim can be further demonstrated.

The distributions of bounding box size for four categories are shown in Figure~\ref{fig:size_dis}. We can find SV2V-RSim includes object annotations with more defined and variable sizes. The differentiation in motor vehicle sizes is mainly reflected in the variety of different vehicle models, while the differentiation in people is seen in their various postures and conditions, such as individuals riding motorcycles or those in wheelchairs, etc.

Figure~\ref{fig:distance} shows that there are more motor vehicles within the 0-5 meter range, while the annotation quantities for the other distance intervals are more evenly distributed across the four categories. This indicates a denser distribution near the ego vehicle, with objects from the four categories also distributed fairly evenly across the remaining distance ranges. 

From the Figure~\ref{fig:keyframe}, it can be observed that the four categories of annotations (Motor Vehicles, Non-Motorized Vehicles, Person, Static Objects) in the key frames are adequately represented. The balanced distribution of annotation counts covers a range of scenarios, from simple to complex.

\begin{figure*}[t]
    \centering
    \begin{minipage}{0.23\textwidth}
        \centering
        \includegraphics[width=\linewidth]{images/data_analysis/motor.png}
        \subcaption{Motor Vehicle}
    \end{minipage}\hspace{0.01cm} 
    \begin{minipage}{0.23\textwidth}
        \centering
        \includegraphics[width=\linewidth]{images/data_analysis/non_motorized.png}
        \subcaption{Non-motor Vehicle}
    \end{minipage}\hspace{0.01cm} 
    \begin{minipage}{0.23\textwidth}
        \centering
        \includegraphics[width=\linewidth]{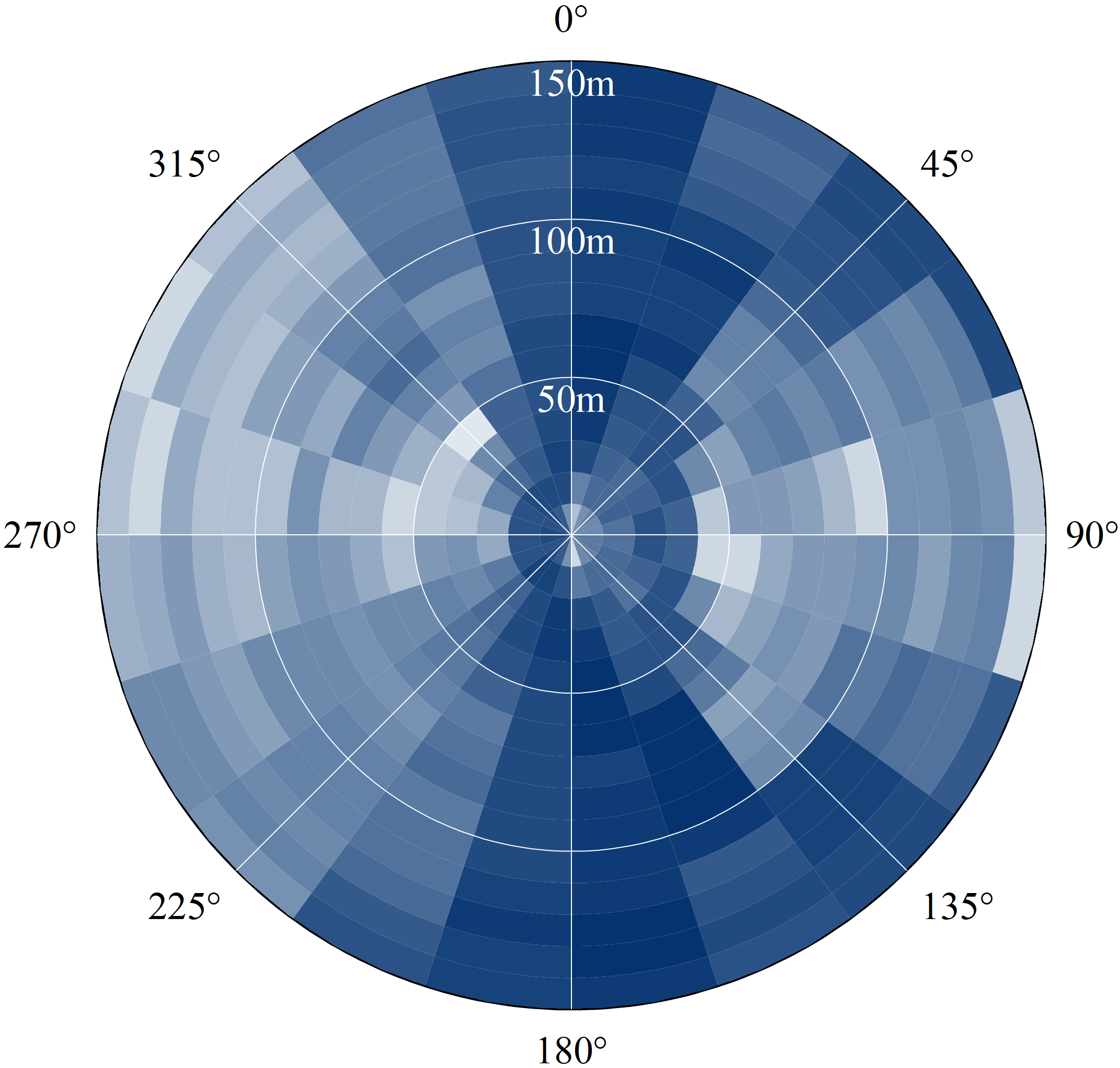}
        \subcaption{Person}
    \end{minipage}\hspace{0.01cm} 
    \begin{minipage}{0.23\textwidth}
        \centering
        \includegraphics[width=\linewidth]{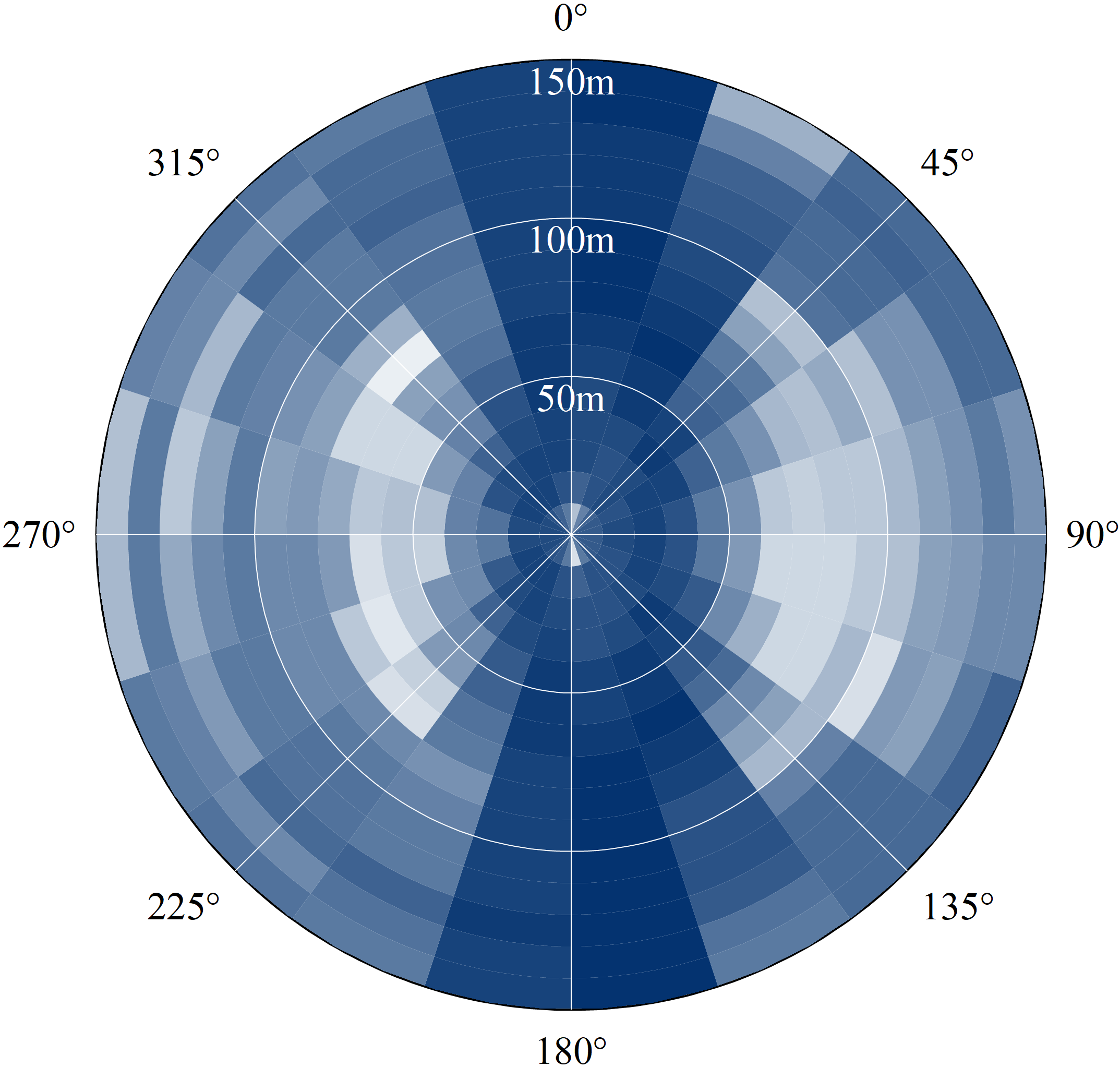}
        \subcaption{Static Object}
    \end{minipage}
    \caption{Polar log-scaled density map of box annotations for four categories, where the radial axis represents the distance from the ego car in meters, and the angular axis corresponds to the yaw angle relative to the ego car. The darker the color, the higher the annotation count in that area.}
    \label{fig:pos_dis2}
\end{figure*}

Figure~\ref{fig:pos_dis2} provides a more detailed radar chart of object distributions. Each category is represented across multiple angular ranges and varying densities. This highlights the dataset’s strength in distributional diversity.

\subsection{Image Quality Assessment (IQA) Metrics}
In this section, we provide a detailed introduction to the IQA evaluation metrics used, thereby illustrating the implications of the advantages reflected in the experimental results.

\textbf{MANIQA}(Masked Attention Network for Image Quality Assessment) uses attention mechanisms and masking strategies to focus on visually important regions, enabling more accurate predictions of perceived image quality. A higher MANIQA score indicates better perceived image quality, reflecting a closer alignment with human visual perception.

\textbf{MUSIQ}(Multi-scale Image Quality) leverages transformer-based models to capture image quality across multiple spatial scales. It effectively models both global and local visual features to predict human-perceived image quality. Similarly, a higher score indicates a closer alignment with human perception of real and high-quality images.

\textbf{NIQE}(Natural Image Quality Evaluator) and \textbf{ILNIQE}(Integrated Local NIQE) are no-reference image quality assessment metrics that rely on statistical models of natural scenes. NIQE evaluates image quality by measuring deviations from statistical regularities observed in high-quality natural images. ILNIQE extends NIQE by integrating local and global features, allowing for more accurate quality assessment, especially in images with non-uniform distortions. Lower NIQE and ILNIQE scores indicate better image quality, as they reflect smaller deviations from the statistical characteristics of high-quality natural images.

\subsection{Experiments Details}
In this section, we offer a more detailed explanation of the baseline methods used in our experiments and our SVA module.

The experiments are conducted on an NVIDIA 4090 GPU to assess the performance of the proposed method across various scenarios. The dataset is split into a 7:3 ratio for training and testing. During training, we set the number of epochs to 30, using the Adam optimizer with an initial learning rate of 0.0002. This learning rate is chosen to ensure stable convergence during training, preventing premature convergence to suboptimal solutions.

For evaluation in cooperative perception scenarios, we define the communication range of the Ego vehicle to be 70 meters. Specifically, all objects within this 70-meter range from the Ego vehicle are considered for the cooperative perception calculation. This setup mimics real-world interactions and collaborative perception between autonomous vehicles within an urban road environment. The 3D detection evaluation is primarily conducted using point cloud data, which allows for precise perception and object recognition in complex environments, especially in dynamic traffic conditions. Moreover, to better generalize the model to real-world applications, we later extend the evaluation to the real-world dataset to assess the performance and adaptability of the benchmark in real-world settings and the reality and genelization of our dataset.

To ensure the reliability and validity of the results, we employ several performance metrics, including the detection results of different categories, to measure the effectiveness of our method. All experiments are carried out on a consistent hardware platform, ensuring the comparability and stability of the results.
\label{sec:experiment_detail}
\subsubsection{Baseline Methods}
In this section, we introduce our backbone and anchor setup in the following two parts:

\indent \textbf{PointPillars backbone.} In all experiments, we configure the PointPillars backbone with a voxel resolution of 0.4 meters in both the x and y directions. The maximum number of points per voxel is set to 32, while the total number of voxels is capped at 32,000.

\textbf{Anchor Setup.} To better accommodate multi-class object detection, we set different anchor sizes for different categories. Specifically, the anchor dimensions are defined as follows: length ($l$) is set to $[4.73, 1.81, 0.96]$, width ($w$) to $[2.08, 0.84, 0.48]$, and height ($h$) to $[1.77, 1.77, 1.2]$ for different object categories.

\subsubsection{Sim2Real Experiments}
In this section, we provide additional details of Sim2Real experiments.

\indent \textbf{Comparison with other simulation data. }For the model, we use the same configuration as the baseline. Additionally, since OPV2V only contains annotations for Cars, we ensure fairness in comparison by removing annotations for all other categories when training with SV2V-RSim. Finally, when testing on the real-world dataset V2XReal, we also exclude other categories.

\textbf{Pretraining with SV2V-RSim. }We randomly select 12.5\% of the data from SV2V-RSim as the pretraining dataset and train for 10 epochs, followed by full-scale training on the complete V2XReal dataset. The improved detection results not only demonstrate the effectiveness of high-quality simulation data in real-world scenarios but also validate the success of SV2V-RSim’s domain gap reduction modules.

